\pdfoutput=1
\documentclass[11pt]{article}

\usepackage[]{naacl2021}

\usepackage{times}
\usepackage{latexsym}

\usepackage[T1]{fontenc}
\usepackage[utf8]{inputenc}

\usepackage{microtype}

\usepackage{multirow}
\usepackage{booktabs}

\usepackage[ruled,linesnumbered,noend]{algorithm2e}
\usepackage{enumitem}
\usepackage{subfigure,xspace}
\usepackage{paralist}
\usepackage{graphicx}
\usepackage{amsmath}

\usepackage{longtable}

\usepackage{hyperref}
\usepackage{url}
\usepackage{comment}
\usepackage[flushleft]{threeparttable}
\usepackage{graphicx}
\usepackage{amsmath}
\usepackage{todonotes}

\usepackage{textcomp}
\usepackage{amsmath}

\usepackage{amssymb}% http://ctan.org/pkg/amssymb
\usepackage{pifont}% http://ctan.org/pkg/pifont
\newcommand{\cmark}{\ding{51}}%
\newcommand{\xmark}{\ding{55}}%

\newcommand{\mym}{SEAL-Shap\xspace}

\newcommand{\corpora}{domains/languages\xspace}

\title{Evaluating the Values of Sources in Transfer Learning}

\author{Md Rizwan Parvez \\
  University of California Los Angeles \\
  \texttt{rizwan@cs.ucla.edu} \\\And
  Kai-Wei Chang \\
  University of California Los Angeles \\
  \texttt{kwchang@cs.ucla.edu} \\}

\date{}

\begin{document}
\maketitle
\begin{abstract}

Transfer learning that adapts a model trained on data-rich sources to low-resource targets has been widely applied 
in natural language processing (NLP).
However, when training a transfer model over multiple sources, not every source  is equally useful for the  target. 
To better transfer a model, it is essential to understand the values of the sources.
In this paper, we develop \mym, an efficient source valuation framework for quantifying the usefulness of the sources (e.g., \corpora) in transfer learning based on the Shapley value method. 
Experiments and comprehensive analyses on both cross-domain and cross-lingual transfers demonstrate
that our framework is not only effective in choosing useful transfer sources but also the source values  
match the intuitive source-target similarity.

%interpretable and  generalized to accommodate different NLP transfer settings.
\end{abstract}

\section{Introduction}
\label{sec:intro}

Transfer learning has been widely used in learning models for 
%is a machine learning method that improves the performance of a 
low-resource scenarios  by leveraging the supervision provided in data-rich source corpora. It has been applied to NLP tasks in various settings including  domain adaptation~\cite{blitzer-etal-2007-biographies,ruder-plank-2017-learning},  cross-lingual transfer~\cite{tackstrom-etal-2013-token,mbert}, and task transfer~\cite{liu-etal-2019-linguistic, vu-etal-2020-exploring}.

% applications, including document classification , NLI \cite{Conneau-lample2018word, xnli, artetxe2019massively},  POS tagging \cite{ ruder-plank-2017-learning}, dependency parsing \cite{tackstrom-etal-2012-cross, guo2015cross, on-difficulties}.

 \begin{figure}[t]
\vspace{-0.2cm}
\hspace{-0.25cm}
\centering
\includegraphics[width=\linewidth]{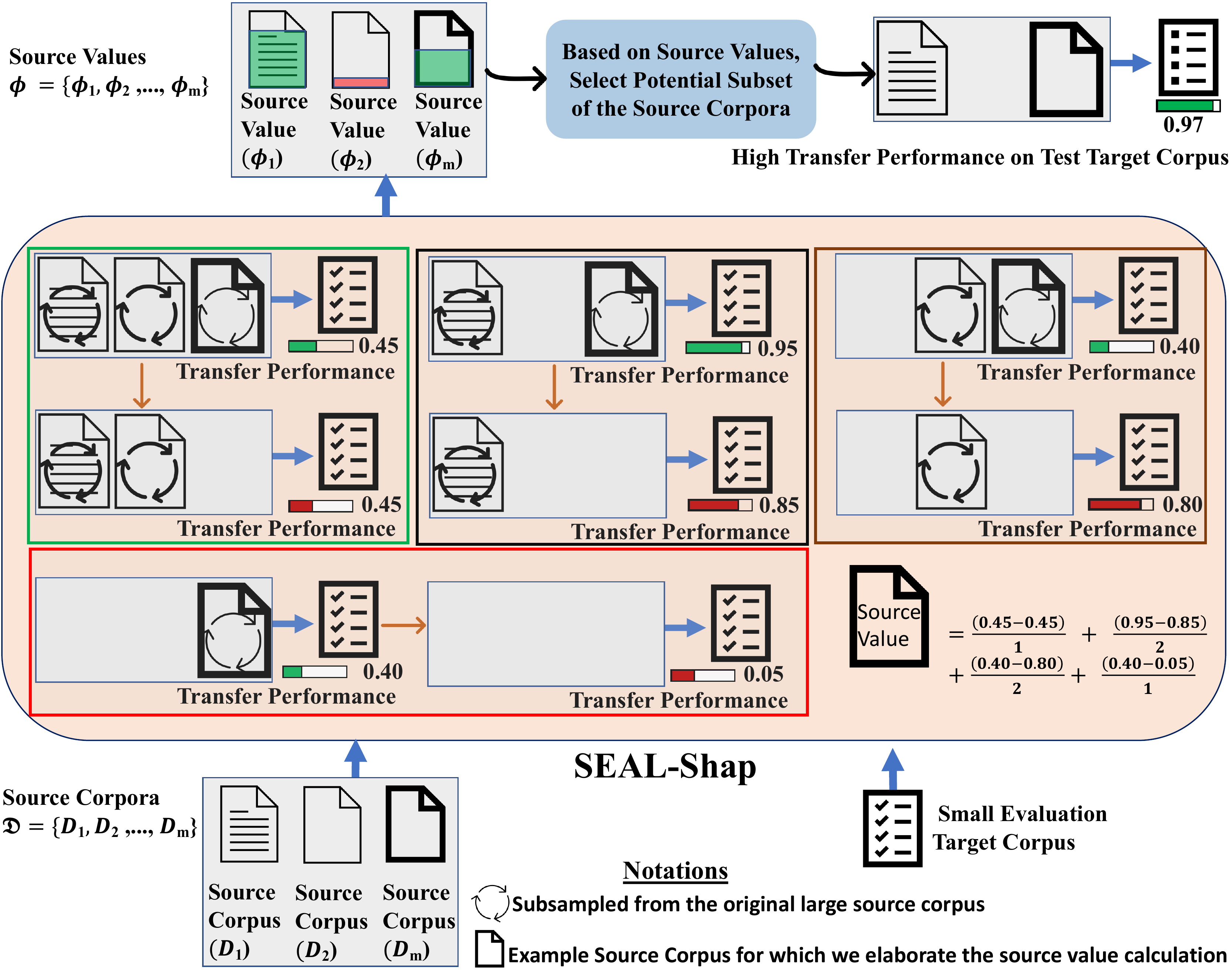}
 \vspace{-5pt}
\caption{
% \vspace{1.5cm}
{\mym estimates the value of each source corpus by the average marginal contribution of that particular source corpus to every possible subset of the source corpora. Each block inside \mym denotes a possible subset and the marginal contribution is derived by the difference of transfer results while trained with and without the corresponding source.  Based on the source values, we select a subset of source corpora  that achieves high transfer accuracy. 
% We also use \mym values as the training labels for a source ranker that is solely based on the data statistics and linguistic features of the source corpora.
}
\vspace{-20pt}
}
  \label{fig:SEAL-SHAP}
\end{figure}

 A common transfer learning setting is to train a model on a set of sources and then evaluate it on the corresponding target~\cite{boosting_from_multi_source, multi-source}.\footnote{In this paper, we focus on two transfer learning scenarios: 1) cross-lingual and 2) cross-domain.  We train a model on a set of source corpora and evaluate on a target corpus where each ``corpus'' refers to the corresponding domain or language.} 
However, not every source corpus contributes equally to the transfer model. Some of them may even cause a performance drop \cite{james-data-shapley, neubig-choosing}.  Therefore, it is essential to understand the value of each source in the transfer learning  not only to achieve a good transfer performance but also for analyzing the source-target relationships.  
%for any target task, understanding the value of each of the source tasks in the transfer learning setting is important. 
% Therefore, a key question arises: how to determine and interpret the value of source tasks in the transfer learning?
% Yet, the majority of existing data selection methods focus only on   high transfer accuracy without realizing the values of a source corpus properly.   

% However,
Nonetheless,
determining the value of a source corpus 
% in the transfer learning 
is challenging as it is affected by many factors, including the quality of the 
source data, the amount of the source data, and the difference between source and target  at lexical, syntax and semantics levels~\cite{on-difficulties,  neubig-choosing}. 
% Consequently, directly evaluating the value of a source corpus is difficult.  Hence,  we consider a data-driven approach
% to derive a data valuation framework
% to understand the value of each source corpus in a generic multi-source transfer learning .
% setting.
% which we then leverage to select a potential subset of source corpora that gains high transfer accuracy. 
The current source valuation or ranking methods are often based on single source transfer performance \cite{mcdonald2011multi, neubig-choosing, vu-etal-2020-exploring} or leave-one-out approaches \cite{ tommasi2009more, DBLP:journals/corr/LiMJ16a, feng2018pathologies, rahimi2019massively}. 
% These approaches evaluates the source values only for one combination either with no other sources or with all other sources. They do not consider the contribution of a source 
% to the other possible subsets of the source corpora.
They do not consider the combinations of the sources. 
Consequently, they may identify the best single source corpus effectively but their top-$k$ ranked source corpora may achieve limited gain in transfer results.
% For example, on Universal Dependencies Treebanks (v2.2)
% % \footnote{\url{github.com/flairNLP/flair/blob/master/resources/docs/TUTORIAL_6_CORPUS.md}} 
% \cite{ud22}, using the top-3 sources predicted by such methods, the zero-shot cross-lingual (here transfer ``corpus'' refers to transfer language) POS tagging accuracy for target English is $\sim$2\% less than our best result (See Fig \ref{fig:prelim-top-3}a).  
% the performance of a transfer model trained either only on the corresponding source or  unaware of performance or drop when trained with a source corpora.

In this paper, we introduce \emph{\mym (Source sElection for trAnsfer Learning via Shapley value)}, a  source valuation framework\footnote{Our source codes  are available at \url{https://github.com/rizwan09/NLPDV/}} (see Fig \ref{fig:SEAL-SHAP})
based on the Shapley value~\cite{shapley1952value, roth1988shapley} in cooperative game theory. \mym adopts the notion of Shapely value to understand the contribution of each source by computing the approximate average marginal contribution of that particular source to every possible subset of the sources.

Shapley value is a unique contribution distribution scheme that satisfies the necessary conditions for data valuation like fairness and additivity \cite{dubey1975uniqueness, knn-shapley-vldb, knn-shapley-arxiv}. As many model explanation methods  including Shapley value are computationally costly \cite{van2021tractability}, in a different context of features and data valuation in machine learning, \citet{james-data-shapley} propose to use an approximate Shapley value to estimate the feature or data values. 

  However, the existing approximation methods for estimating Shapley values are not scalable for NLP applications.
% When the number of sources is high, they require to re-train the corresponding model many times. Besides, 
NLP models are often large (e.g., BERT \cite{devlin2019bert}) and NLP transfer learning usually assumes a large amount of source data. To deal with the scalability issue, we propose a new sampling scheme, a truncation method, and a caching mechanism to efficiently approximate the source Shapley values.

We evaluate the effectiveness of \mym  under various applications  in
quantifying 
the usefulness of the source corpora 
% regardless of an evaluation corpus,
% . Second, we present 
% it can effectively be used 
% to select 
and in selecting
potential transfer sources. We consider two settings of source valuation or selection: (1) where a small target  corpus is available; and (2) where we only have access to the linguistic or statistical features of the target, such as language distance to the sources,  typological properties, lexical overlap etc. For the first setting, we use the small target data as the validation set to measure the values of the sources w.r.t the target. For the second setting, we follow \citet{neubig-choosing} to train a source ranker based on \mym and the available features.

We conduct extensive experiments  in both (zero-shot) cross-lingual and cross-domain  transfer settings on three NLP tasks, including POS tagging, sentiment analysis, and natural language inference (NLI) with different model architectures (BERT and BiLSTM). 
% We also show that our source values are meaningful and follows the intuitive source-target relationships.
In a case study,
% source languages 
% in 
on
the cross-lingual transfer
learning,
% and
we
exhibit that the  source language values are
  correlated with the language family and language distance\textemdash indicating that our source values are meaningful and follow the intuitive source-target relationships.
% Lastly, we  analyze \mym in details. 
Lastly, we analyze  the approximation correctness and the run-time improvement of our source valuation framework \mym.

\section{Source Valuation Framework}
\label{sec:framework}
  We propose \mym, a source valuation framework.
%   However, it can also be leveraged to achieve performance gain by choosing the useful source domains/languages in  transfer learning.
  We start with  the setting where we have only one target and multiple sources. We denote the target corpus by $V$ and the corresponding set of source corpora by $\mathcal{D}= \{D_1,\cdots, D_m\}$.  Our goal is  to quantify the value $\Phi_{j}$ of each source corpus $D_j$ to the transfer performance on $V$ and explain model behaviors. Once the source  values are measured, we can then develop a method to select either all the sources or a subset of sources (i.e., $\subseteq \mathcal{D}$) that realizes a good transfer accuracy on $V$.  
% The source corpora values 
% can then be used to select source corpora for transfer learning. 
Below,  we first review the data Shapley value and its adaptation for transfer learning. Then, we describe how \mym efficiently quantifies $\Phi_{j}$ and how to use
it to select a subset of sources for model transfer.

%We further extend \mym to simultaneously valuate the sources w.r.t a set of target corpora. 

%we discuss how to use \mym and estimates source values when no evaluation corpus is available at inference time. Finally we discuss how to select a potential subset of $\mathcal{D}$. 

%describe how to compute the source-task value by 
%approximated data Shapley value for the source tasks by adapting an existing approach TMC-Shap \cite{james-data-shapley} in \ref{sec:tmc-shap-james-section}. Then in Section 
%\ref{sec:ftmc-shap}, we develop \mym for NLP transfer learning setting. 
%Finally, in Section \ref{sec:source-selection}, we discuss how to select the potential source tasks based on the \mym value.
% \subsection{Overview of Our Framework}

\subsection{Background: Data Shapley Value }
\label{sec:tmc-shap-james-section}
Shapley value is designed to measure individual contributions in collaborative game theory and has been adapted for data valuation in  machine learning~\cite{james-data-shapley, knn-shapley-vldb, knn-shapley-arxiv}. 
In the transfer learning setting, on a target corpus $V$, let $Score(C_{\Omega}, V)$ represent the transfer performance 
of a model $C$ trained on a set of source corpora $\Omega$.\footnote{ 
In this paper, we consider a model trained on the union of the source data and the loss function for training the model is aggregated from the loss functions defined on each source. However, our approach is agnostic to how the model is trained and can be integrated with other training strategies.}
The Shapley value 
%given a set of source tasks $\mathcal{D}= \{D_1,\cdots, D_m\}$, let $C_D$ be a model trained on a set of source tasks $D$  and for a target task $V_k$, the  Shapley value of a source task $D_j$ by $\Phi_{j,k}$. 
%Given a performance measure $Score(C,V_k)$, the Shapley value 
$\Phi_{j}$ is defined as the average marginal contribution of a  source corpus $D_j$ to every possible subsets of corpora $\mathcal{D}$:
\begin{equation*}
\small
\label{eq:exact}
     \frac{1}{m} \sum_{\Omega \subseteq \mathcal{D}-D_j}\!\! \frac{Score(C_{\Omega \cup D_j}, V) \!-\! Score(C_\Omega, V)}{\binom{m-1}{\lvert \Omega \rvert}}.
\end{equation*}

% We propose our source value approximation framework \mym in Algorithm \ref{algo-1}. 

\paragraph{ TMC-Shap for Transfer Learning:}
Computing the exact source-corpus Shapley value, 
% in the  
% Eq \eqref{eq:exact}
described above,
is computationally difficult as it involves evaluating the performances of the transfer models trained on all the  possible  combinations of the source corpora. Hence, \citet{james-data-shapley} propose to approximate the evaluation by a truncated Monte Carlo method. 
% The exact Shapley value being computationally exponential, following \citet{shap-tmc-1, shap-tmc-2, shap-tmc-3}, \citet{james-data-shapley} develops a Truncated Monte Carlo approximation (TMC-Shap) method and shows that it efficiently estimates data Shapley values and suitable for neural networks. 
%Here we  adopt the TMC-Shap \citet{james-data-shapley} method for transfer learning.
Given the target corpus $V$ and a set of source corpora $\mathcal{D}$,
for each epoch, a source training data set $\Omega\subseteq \mathcal{D}$ is maintained and a random permutation $\pi$ on $\mathcal{D}$ is performed
% { \color{red} { 
(corresponds to line 6 in Algorithm \ref{algo-1} which is discussed in Sec \ref{sec:ftmc-shap}).
% }}
Then it loops over every source corpus $\pi_j$ in the ordered list $\pi$ and
%add it to the training set of source tasks $\Omega$
%\footnote{Note that $\Omega$ %\subseteq \mathcal{D}$; $D_\pi$ is ordered, and $d$ is order-free.} 
compute its marginal contribution by evaluating how much the performance improves by adding $\pi_j$ to $\Omega$: $Score(C_{\Omega \cup \pi_j}, V) - Score(C_\Omega, V)$.
These processes are repeated multiple rounds and 
%However, this marginal contribution of the corresponding source task is one monte-carlo sample and hence we repeat over multiple epochs and 
 the average of all marginal contributions associated with a particular source corpus is taken as its approximate Shapley value
% { \color{red} { 
(line 18 in Algorithm \ref{algo-1}).
% }}
When the size of $\Omega$ increase, the marginal contribution of adding a new source corpus becomes smaller. Therefore, 
to reduce the computation, \citet{james-data-shapley} propose to truncate the computations at each epoch when the marginal contribution of adding a new source  $\pi_j$ is smaller than a user defined  threshold {\it Tolerance}
% { \color{red} {
(line 10-11, 18 in Algorithm \ref{algo-1}).\footnote{Setting {\it Tolerance} to 0 turns off the truncation.}

% }}

%Because with the increase of $\Omega$ for the next source task $D_{\pi, j}$, the marginal contribution  becomes smaller. Therefore in each epoch $t$, we truncate the computations once the marginal contribution of a source task becomes small and approximate the marginal contribution of the following source tasks with zero (line 18 in Algorithm \ref{algo-1}). 

% \vspace{-5pt}
\subsection{\mym} 
\label{sec:ftmc-shap}
Despite that TMC-Shap improves the running time, it is still unrealistic to use it  in our setting where both source data and model are large. 
For example, in cross-lingual POS tagging on  Universal Dependencies Treebanks, on average, it takes more than 200 hours to estimate the values of 30 source languages with multi-lingual BERT (See Sec \ref{sec:ablation}). Therefore, in the following, we propose three techniques to further speed-up the evaluation process.
% In the following, we introduce \mym to adopt TMC-Shap for handling large scale problems. 

\noindent{\bf Stratified Sampling} 
When computing the marginal contributions, training a model $C$ on the entire training set $\Omega$ is computationally expensive. Based on extensive experiments, when computing these marginal contributions, we find that we do not need the performance difference of models trained with the entire training sets. 
% Instead,  a proportionate performance difference is sufficient.
For a reasonably large source corpus,  20-30\% samples\footnote{Higher sampling rate typically leads to better approximation but are expensive in run-time.}  in each source achieve lower but representative performance difference, in general. Therefore, we sample a subset of instances to evaluate the marginal contributions.
To address computational limitation and scale to large data, sampling techniques have been widely discussed \cite{l2017machine}.
% ~\cite{DBLP:conf/icml/DauphinGB11}.
In particular, we employ a stratified sampling~\cite{neyman1992two} to generate a subset $\mathcal{T}$ from $\Omega$ by sampling training instances from each source corpus $\Omega_x$ with a user defined  sample rate $\eta$. Then, we train the model on $\mathcal{T}$ (line 14-15 in Algorithm \ref{algo-1}).
 The quantitative effectiveness of this technique is discussed in Sec \ref{sec:ablation} and the impact of different sampling rates are presented in Fig \ref{fig:how-good-3datasets}. 
%With enough number of epochs, the 
%As more number of samples are often better than once, we re-sample each source task whenever the classifier is need to be tr
\begin{algorithm}[t]
%\SetAlgoLined
\DontPrintSemicolon
\small
%\Indm
%\crzwn{$S_w$ and S is undefined. Should I change it as previous?}\\

\KwIn{Source corpora $ \mathcal{D} = \{D_1,\cdots,D_m\}$, target corpus $V$, Random sampler  $\mathcal{S}$,   sample size $\eta$, num of epochs $nepoch$, and Classifier $C$}
\KwOut{Source-corpora Shapley values $\{\Phi_{1}...,\Phi_{m}\}$ }
%\Indp
% \vspace{3pt}
%  $C_1 \gets$ Train a \textit{classifier} $C$ on $\mathcal{X}$\;  
 {\bf Initialize:}  Score cache $S \gets  \{\}$, source  Shapley values $\Phi_{x} \gets 0$ for $x$ = $1\ldots m$, and epoch
$ t \gets  0$ \;
    ${\mathcal{D}_{samp}} \gets \{\mathcal{S}(D_x, \eta), \forall D_x \in \mathcal{D}\}$ \;
    $ C_{{\mathcal{D}_{samp}}} \gets$ Train $C$ on ${\mathcal{D}_{samp}}$ \;
    
    \While{Converge or $t < nepoch$} {    
        
        $t \gets t+1 $\;
        $\pi: $ Random permutation of $\mathcal{D}$\;
        $v_{0} \gets \rho$ \;

       \For{$j \in \{1, \cdots m \} $}
       {   
            $\Omega \gets \{\pi_{1},\cdots, \pi_{j}\}$ \;
        
            \uIf{$\lvert$ Score($C_{{\mathcal{D}_{samp}}} , V)$ -                 $v_{j-1}\rvert  <$  Tolerance}
                {   
                    $v_{j} \gets v_{j-1}$\;
                }
            \uElse{

                \uIf{$\Omega \notin S$}
                    {
                        
                         $\mathcal{T} \gets \{\mathcal{S}(\Omega_x, \eta), \forall \Omega_x \in \Omega\}$ \;
               
                         $C_j \gets $ Train $C$ on $\mathcal{T}$\;
                
                        Insert $\Omega$ into $S$ with $S_\Omega \gets $ Score$(C_j , V)$\;
                        
                    }
                 $v_{j} \gets S_\Omega$\;
            }

            $\Phi_{\pi_j} \gets \frac{t-1}{t} \Phi_{\pi_j} + \frac{1}{t}
            ( v_{j} - v_{j-1}) $\;
            
        }

   }
 \caption{{\small \bf \mym}}
 \label{algo-1}
%  \vspace{-5pt}
\end{algorithm}
\\
\noindent{\bf Truncation}  
%As noted in \citet{smaller-tail2, smaller-tail1}, the marginal contribution of a new source task $D_j$ often decreases while the size of $\Omega$ grows.
%with the increase of the size of the training data set $\Omega$ and the highest marginal contribution is usually  observed at the beginning of each epoch when training data set is empty. 
%As discussed in Section \ref{sec:tmc-shap-james-section}, in each epoch, 
As discussed in Sec \ref{sec:tmc-shap-james-section}, at each epoch, \citet{james-data-shapley} truncate the computations once a marginal contribution becomes small 
% and assume that the contribution of the remaining corpora is 0 (line 10-11, 18 in Algorithm \ref{algo-1}).
when looping over the  ordered list $\pi$ of that corresponding epoch, typically for the last few sources in $\pi$.
On the other hand, at the beginning of each epoch, when computing the marginal contribution by adding the first source corpus $\pi_1$ into an empty $\Omega$, the contribution is computed by the performance gap between a model trained on $\pi_1$ and a random baseline model without any training. Usually, the performance of a random model ($v_0$) is   low and hence, the marginal contribution is high in the first step, in general. As this scale of marginal contributions at the first step is drastically different from later steps, it leads TMC-Shap to converge slowly. Hence, to restrict the variance of the marginal contributions,
 we down weight the marginal contributions of the first step by setting $v_0$ = $\rho$, where $\rho$ is a hyper-parameter\footnote{Typically a factor of  the performance achieved when using only one source, or all the sources together} indicating the baseline performance of a model (line 7, 18 in Algorithm \ref{algo-1}).

%then approximate the marginal contributions of the following source tasks with zero and starts the next epoch with with a new permutation 
%(line 18 in Algorithm \ref{algo-1}).
%Consequently it takes more epochs to get a good approximation.  To facilitate a good approximation given a certain number of epochs $nepoch$, along with this truncation when $\Omega$ becomes large, we also truncate (i.e., down-weight) the high marginal contribution when $\Omega$ is empty. To do so, we empirically choose a high initial performance measure  (line 9 in Algorithm \ref{algo-1}) and for the first source task  $D_{\pi,1}$ in the permutation $D_\pi$, we compute the marginal contribution based on how much it improves the performance measure from that high initial score
%instead of a very low performance measure of a randomly initialized classifier (i.e., $Score(C_{\emptyset}, V_k)$).

\noindent{\bf Caching} When computing the source Shapley values, we have to repeatedly evaluate the performance of the model on different subsets of source corpora. 
%are indeed considering the performance difference of the model trained with different training sets (i.e.,  different subsets of the source tasks).  Now, 
%When evaluating the performance difference with different subsets, 
Sometimes, we may encounter subsets that we have evaluated before. For example, consider a set of source corpora $\mathcal{D}=\{D_1,D_2,D_3\}$ and we evaluate their Shapley values through two permutations: $\pi_1 = [D_3,D_1,D_2]$, and $\pi_2 = [D_1,D_3,D_2]$. 
When we compute the marginal contribution of the last source corpus $D_2$, in both cases the training set $\Omega = \{D_1,D_3\}$. 
%is common in both $\pi_1$ and $\pi_2$. 
That is, if we cache the result of $Score(C_{D_1\cup D_3})$, then we can reuse the scores. We implement this cache mechanism in line 1, 13, 16, 17 in Algorithm \ref{algo-1}. With these optimization techniques, we improve the computation time by  about $2$x (see Sec \ref{sec:ablation}). This enables us to apply this techniques in NLP transfer learning.

Note that whenever an $\Omega$ causes a cache miss, for each source $\Omega_x$, as discussed above in this Section, we sample a new set of instances (line 13-14 in Algorithm-1). Thus, given a reasonably large number of epochs, our approach performs sampling for a large number of times and in aggregation, it evaluates a wide number of samples in each source.

%In such a case, there is no need to retrain the model and reevaluate again. Therefore, once we train the classifier $C$ on a new training set $\Omega$, we cache the performance measure. For any further iteration in the following epochs, if we see the same training set, we reuse the cached performance (line 1, 12, 16, 21 in Algorithm \ref{algo-1}). 

%\paragraph{Multi-target valuation:} 
%when considering evaluating Shapley values for multiple target tasks at the same time, we can reuse the model trained on 
%Many practical transfer learning problems have multiple target tasks with a set of common source tasks. In such cases, \mym may encounter same training set in multiple target tasks. For example, lets consider that for  two target tasks $V_1$, $V_2$, the set of the source tasks are $\mathcal{D}_1 = \{1,2,3\}$, and $\mathcal{D}_2 = \{1,3,4\}$.  Then, the training set $d = \{1,3\}$ is common in $\mathcal{D}_1$ and $\mathcal{D}_2$. 
% Now for the two corresponding permutations $[1,3,2]$, and $[3,1,4]$ for $\mathcal{D}_1$ and $\mathcal{D}_2$ respectively, the training set $d = \{1,3\}$ is common. 
%Therefore, to avoid repetitive training on the same training set $\Omega$, for the first encounter of $\Omega$, \mym trains the classifier $C$ on $d$ regardless of a specific target task $V_k$ (i.e, training without any hyper-parameter tuning). And then instead of evaluating $C_\Omega$ only on $V_k$, we evaluate on all the target tasks and cache all the  the performance measure (line 15-17 in Algorithm \ref{algo-1}). \mym is summarized  in Algorithm \ref{algo-1}.

\subsection{\mym for Multiple Targets } 
% \paragraph{\mym for multiple Targets: } 
\label{sec:ftmc-shap-multi-target}
Many applications require to evaluate the values of a set of sources with respect to a set of targets. 
%Here, we discuss how to reduce the run time where there are multiple targets with overlapping sources. 
For example, under the zero-shot transfer learning setting, we assume a model is purely trained on the source corpora without using any target data. Consequently, then the same trained model can be evaluated on multiple target corpora. 
With this intuition, 
% when a new model is trained, we cache it and use it to 
% evaluate on all target corpora to reduce the redundant computations. 
% Therefore, to reduce the redundant computations while computing the Shapley values for any target corpus, 
% \mym caches the transfer performances on all other target corpora also. 
% Whenever 
whenever the model is trained on a new training set $\Omega$, \mym evaluates it on all the target corpora and caches all of them accordingly.
% for reducing redundant computations when there are multiple targets. 

\subsection{Source Values without Evaluation Corpus}
\label{sec:direct-source-value}
In the previous discussions above, we assume a small annotated target corpus is available and can be used to evaluate the transfer performances. 
However, in some scenarios,  only some linguistic or statistical features of the sources and targets, such as language distance and word overlap, are available. 
\citet{neubig-choosing} show that by using these features, we can train a ranker to sort the sources to unknown targets by predicting their value. 
%the value of sources to unknown targets. 
In the following, we extend their ranker by incorporating it with \mym. 
%in some scenarios, an annotated dataset in target $V$ may not be accessible. Here, 
%In this case, we compute the source values in such cases assuming that we have some readily available 

%, we train the fast source ranker in \citet{neubig-choosing} that predicts the source values solely based on these aforementioned features. 

Given the set of training corpora $\mathcal{D}$ and the actual target corpus $V$, we iteratively consider each training corpus $D_j$ as target and the rest $m$-$1$ corpora as the sources. 
% (i.e., $\mathcal{D}_{temp}=\{D_1, \ldots, D_{j-1}, D_{j+1}, \ldots, D_m\}$).
We compute the corresponding source values $\mathcal{Y}^{D_j}_\mathcal{D}=\{\Phi_{D_1}, \ldots  , \Phi_{D_{j-1}}, \Phi_{D_{j+1}}, \ldots , \Phi_{D_m}\}$.  Now, w.r.t the  target $D_j$, the linguistic or statistical features of the source corpora (e.g.,  language distance from the target, lexical overlap between the corresponding source and the target) 
$\mathcal{X}^{D_j}_\mathcal{D} = \{F^j(D_1),\! \ldots,\! F^j(D_{j-1}), F^j(D_{j+1}), \!\ldots,\! F^j(D_m)\}$ where $F^j$ denotes the source feature generator function for the corresponding target $D_j$. This  feature vector of the source corpora ($\mathcal{X}^{D_j}_\mathcal{D}$) is a training input and their  value vector  ($\mathcal{Y}^{D_j}_\mathcal{D}$) is the corresponding training output for the ranker.
We repeat this for each training corpus and generate the respective training inputs and outputs for the ranker.   
% Then we train the ranker using training dataset of $D_j$ as input and its value $\Phi_j$ as label. 
Once trained, for the actual target $V$ and the source corpora $\mathcal{D}$, the ranker can predict the   values of the source corpora $\mathcal{Y}^V_\mathcal{D}$  only based on the  linguistic source features $\mathcal{X}^V_\mathcal{D}$.

%only needs features of the corresponding sources ($F_\mathcal{D}$) as input and  predicts the source values  $\Phi$. 

% We consider the existing ranker \citet{neubig-choosing} that takes a target corpus, the corresponding source corpora as input and predicts the ranking of the source corpora as output. The ranker only considers the readily available features like size of the source corpus, linguistic distance, word overlapping w.r.t the target etc., and therefore at inference time, it does not require any actual transfer models. However, to train the ranker \citet{neubig-choosing} uses the ranking based on the single source transfer performance as annotated labels. We replace this with the ranking based on the \mym values.

\subsection{Source Corpora Selection by \mym}
\label{sec:source-selection}
% For a target $V$, after computing the \mym values of the source corpora $\Phi = \{\Phi_{1}, \cdots, \Phi_{m}\}$,
The source values computed in Sec \ref{sec:ftmc-shap}-\ref{sec:direct-source-value} estimate the usefulness of the corresponding transfer sources and can be used to identify the potential sources which lead to the good transfer performances.   We select the potential source corpora in two ways. (i) Top-$k$: We simply sort the sources based on their values and select the user defined top-$k$ sources. (ii) Threshold: When an annotated evaluation dataset in target corpus $V$ is available, 
% we first use it to compute the \mym values of the sources. Next, 
after computing the source values,
we empirically set a threshold $\theta$ and select each source  that has source value higher than $\theta$. On that evaluation target corpus, we  tune and set $\theta$ for which the corresponding transfer model achieves the best performance. 
% While tuning, if there is no $\theta$ for which the corresponding subset of sources (i.e., $\subset \mathcal{D}$) achieves better result than using all of $\mathcal{D}$, then we select the set of all  sources  $\mathcal{D}$ assuming each source is contributing positively. 

 %achieve a good transfer performance on the target task: $Z_k  = \{ D_j: \delta(\Phi_{j,k} > \theta_k) \forall D_j \in D $ if $D_j \in M_k\},$ where $\delta$ is an
%indicator function.

\begin{figure*}[h]
\setlength{\abovecaptionskip}{-4pt}
\vspace{-0.4cm}
% \hspace{-25pt}
  \centering
    \begin{tabular}{c@{}c@{}}
    \hspace{-0.6cm}
% \vspace{-1.5cm}
  \subfigure[UD Treebank, target: en]{\
        \includegraphics[height=0.25\linewidth, width=0.25\linewidth]{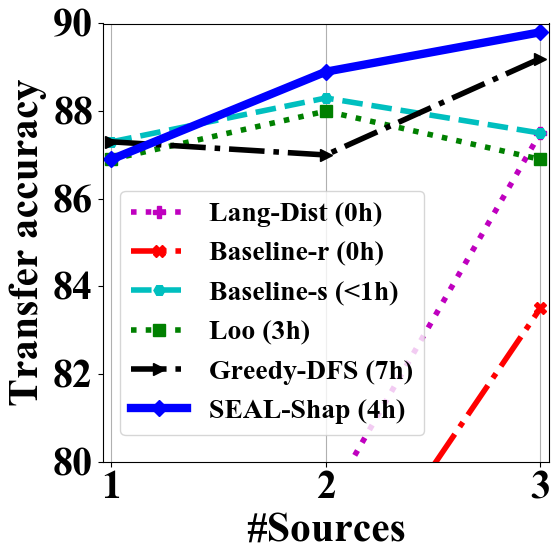}\label{fig:eng-pos-all}
        } 
\hspace{-0.3cm}
 \subfigure[XNLI, target: vi]{\
        \includegraphics[height=0.25\linewidth, width=0.25\linewidth]{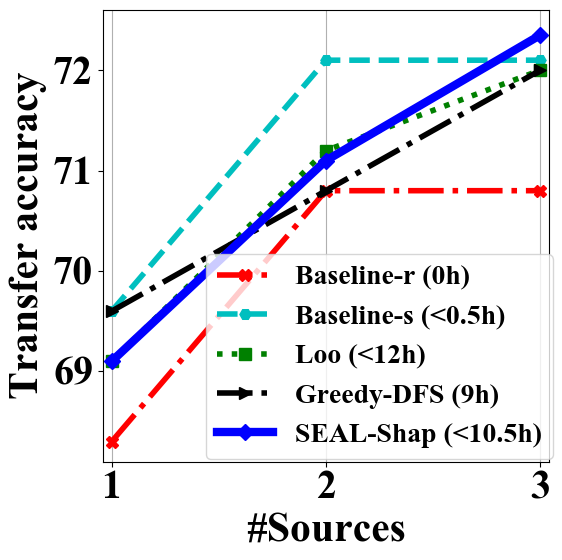}\label{fig:eng-pos-all}
        }
        
\hspace{-0.3cm}
 \subfigure[mtl-dom-senti, target: E]{\
        \includegraphics[height=0.25\linewidth, width=0.25\linewidth]{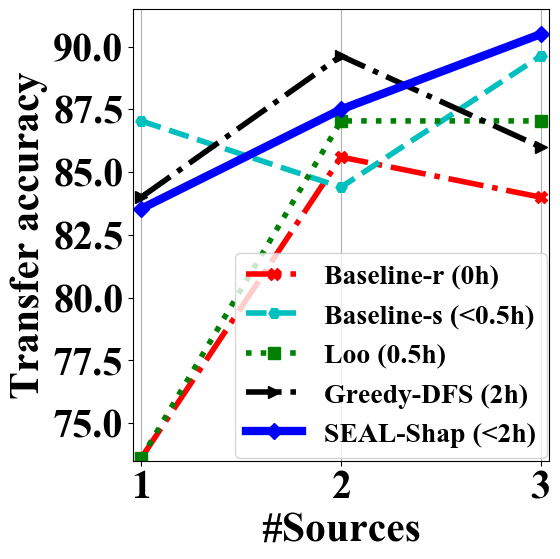}\label{fig:eng-pos-all}
        }

\hspace{-0.3cm}     
    \subfigure[ {{mGLUE, target:MNLI-mm}} ]{\
    \includegraphics[height=0.25\linewidth, width=0.25\linewidth]{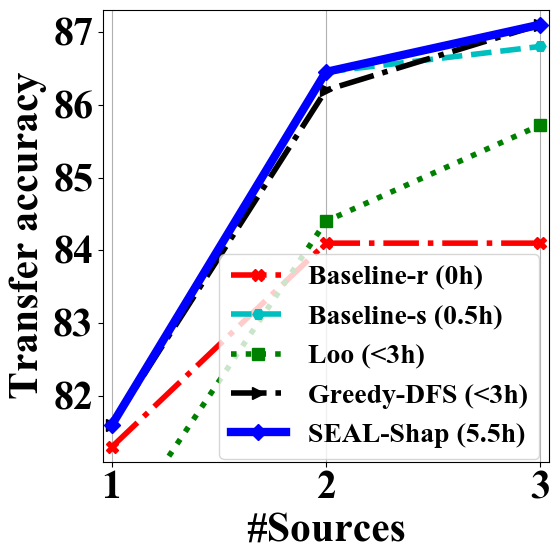}\label{fig:eng-pos-all}
    }
        % &

    \end{tabular}
  \caption{ 
  \footnotesize{Performance, and  run time with up to top-3 sources ranked by different  approaches. (a), (b) denotes cross-lingual and (c), (d) denotes cross-domain transfer. All models have same training configurations (e.g., sample size). All the run times are final except for {\it Greedy DFS} where it increases linearly with top-$k$.  Adding top-2 and top-3 ranked sources, other methods drop their accuracy across the tasks while ours shows a consistent gain in all tasks and achieves the best results with top-3 sources.
%   the 
%   top-3 sources. 
}   
  } 
  \label{fig:prelim-top-3}
\vspace{-0.8em}
\end{figure*}

\section{Experimental Settings}
\label{experiments-ta}
% To evaluate our approach, we conduct experiments on several benchmark datasets both in language and domain transfer settings. In this section, first we discuss the transfer tasks, datasets, and  the settings in our experments. Next, we discuss the zero-shot transfer learning results using our method. Then we show that given a particular task, how the Shapley values of the transfer sources can be an explanation of the  relevance between the source and the target. Finally we present some analysis and ablation study. 
We conduct experiments on \emph{zero-shot cross-lingual and cross-domain transfer} settings. 
Models are trained only on the source languages/domains
 and directly applied in target domains/languages.
%  through the underlying representation models.  
% and then applied to the target language/domain.  
% Evaluating the values of the source language/domains that are very different in size, the marginal contributions of a source may become biased to its size. Hence,
% For an unbiased valuation, we compute source values with the source size. 
% Then, to amplify the performance gain, we report the results using the entire datasets of the selected sources.
% {\color{red}{Note that this may amplify both the benefits of selecting a useful source and harm of selecting an irrelevant source.}} 
% For the evaluation of source values, we consider a small development set in the target language/domain.   
% \subsection{Problems and Datasets}
\noindent{\bf Cross-lingual Datasets } 
We conduct experiments on two popular cross-lingual transfer problems: (i) universal POS tagging on the 
% \href{https://github.com/flairNLP/flair/blob/master/resources/docs/TUTORIAL_6_CORPUS.md}{Universal Dependencies Treebanks (v2.2)}
Universal Dependencies Treebanks
% \footnote{\url{github.com/flairNLP/flair/blob/master/resources/docs/TUTORIAL_6_CORPUS.md}} 
\cite{ud22}.
Following \citet{on-difficulties}, we select 31 languages of 13 different language families (details in Appendix
% A).
\ref{app:langstat}).
% \footnote{Please refer to the supplementary materials for all the appendices of this paper.}
(ii) natural language inference on the XNLI dataset \cite{xnli}, that covers 15 different languages. XNLI task is a 3-way classification task (entailment, neutral, and contradiction).
% , to a pair of input sentences.  
%\ref{app:langstat}
Data statistics are in Appendix 
% R.
\ref{app:datasatat}.
% Table \ref{tab:stat-task}.
% summarises all our datasets.

% \subsection{Classifier and Data Preprocessing}
\noindent{\bf Cross-domain Datasets } We consider three  domain transfer tasks: (i) POS tagging: we use
the SANCL 2012 shared task datasets~\cite{sancl2012} that has six different domains (details in Appendix \ref{app:B}). (ii) Sentiment analysis: we use the multi-domain sentiment datasets~\cite{liu-etal-2017-adversarial} which has several additional domains than the popular \citet{blitzer-etal-2007-biographies} dataset, See Appendix \ref{app:D}. (iii) NLI: we consider a (modified) binary classification (e.g., entailed or not) dataset used in \citet{bert-domain}. It is made upon  modification on GLUE tasks \cite{glue} and has four domains (details in Appendix \ref{app:C}). As GLUE test sets are unavailable, for each target domain, we use the original dev set as the pseudo test set and randomly select 2,000  instances from its training set as the pseudo dev set.  
%Instead of the fact that \mym uses sampling, while computing the source values for any valuation approach we also use sampled datasets. To reason, in an example UD Treebank datasets, the source size of language Czech, and Turkish are 69k, 3.5k respectively. Therefore, using the entire datasets of very different sized sources may lead to a source value biased to its size. For the minimum source size ($\eta$), we sample $\eta$ number of instances from each source (except for cross-doamin NLI task where we set $\eta=20,000$) for all the methods.   

\noindent{\bf Classifier and Preprocessing}
% \vspace{-15pt}
% As for the underlying classifier model, 
For all domain transfer tasks, we use BERT and for all language transfer tasks, we use multi-lingual BERT \cite{devlin2019bert} models except for cross-doman POS tagging where we consider the state-of-the-art BiLSTM based  Flair framework \cite{flair-18}. For BERT models, we use the Transformers implementations in the Huggingface  library \citet{HuggingFacesTS}. For significance test, we use an open-sourced library.\footnote{\url{github.com/neubig/util-scripts/blob/master/paired-bootstrap.py}}  By default, no preprocessing is performed except  tokenization (see Appendix \ref{app:J}). 
% Following \citet{mbert}, for all tasks, we also limit subwords sequence length to 128 to fit in a single GPU. 
% \subsection{Hyper-parameters Tuning: } 

\noindent{\bf Hyper-parameters Tuning}
% All the models are tuned on the target dev corpus. Following \citet{devlin2018bert, mbert},
% For all BERT models, we tune the Adam optimizer with learning rate in $\{2, 3, 5\}\times 10^{-5}$; and batch size $\{16, 32\}$, and number of epochs up to 4.  By default, for a target $V$ and its sources $\mathcal{D}$, number of epochs $nepoch$ in Algorithm \ref{algo-1} is 30, the corresponding threshold \mym value  $\theta$ is chosen in $\{10^{-2}, 10^{-3}, 5\times 10^{-3}\}$, initial scores $\rho$ in $\{0.5, {\textit{ All sources}/2, \textit{ All Sources}}, \mu  \} $  where \textit{ All Sources} is the transfer performance using the  set of all sources $\mathcal{D}$ (see Sec \ref{sec:eval_source-valuation}); $\mu$ is the mean of \textit{ All Sources}, and all the single source transfer performances. Details are in Appendix K.
% For the significance test we use an open-sourced implementation\footnote{\url{github.com/neubig/util-scripts/blob/master/paired-bootstrap.py}}. 
For all BERT models, we tune the learning rate, batch size, and number of epochs.  We also tune the number of epochs $nepoch$ in Algorithm \ref{algo-1}, the threshold \mym value  $\theta$, initial score $\rho$. Details are in Appendix \ref{app:K}.

\section{Results and Discussion}
In the following, we first verify \mym is an effective tool for source valuation. 
% we demonstrate that it is also effective in selecting useful transfer sources in Sec \ref{sec:Evaluating Source Tasks Selection}. 
Then, we evaluate the source values when an evaluation target corpus is unavailable.
 In Sec \ref{sec:interpretablity}, we interpret the relations between sources and targets based on the SEAL-Shap values.   Finally, we analyze our method with comprehensive ablation studies.

\subsection{Evaluating Source Valuation}
\label{sec:eval_source-valuation}

% Here we verify \mym by examining its ability in  selecting potential source corpora.
% locating source corpora that cause performance gain (or drop). %We demonstrate that the source tasks selected by \mym is meaningful such that the corresponding model performance  is higher and hence \mym is able to identify the value of the source tasks correctly. 

We assess our source valuation approach in compare to the following baselines:
(i) {\it Baseline-s}: source values are based on the single source transfer performance.
% This is a brute force method to find the optimal (only one) source and the upper bound of \citet{neubig-choosing}.
% I'm not sure what do you mean by brute force. 
% Also, you haven't introduce the setting of Lin et al thereofre, readers won't be able to understand this sentence if you say it here. Mention this later in the corresponding section if needed.  
(ii) {\it Leave-one-out (LOO)}: source values are based on how much transfer performance we loose if we train the model on all the sources except the corresponding one. 
(iii) {\it Baseline-r}: a random baseline that assigns random values to sources.\footnote{Our experiments with different seeds result in different but similar results.}
% \footnote{Our preliminary results posits that leave-one-out is no better than {\it Baseline-s} as shown in \citet{james-data-shapley}, hence we do not compare with this.}
(iv) {\it Greedy DFS}: the top-1 ranked source is same as that of {\it Baseline-s}. 
% (i.e., $G_1 = \underset{D_j}{\arg\max}$ $Score(C_{D_j}, V)$). 
Next, it selects one of the remaining sources as top-2 that gives the best transfer result along with the top-1 and so on.   
(v) {\it Lang-Dist}: (if available) in reverse order of target-source language distance~\cite{on-difficulties}.\footnote{\citet{on-difficulties} compute the distances from an annotated dependency parse tree based on UD Treebank.}
% In  Fig \ref{fig:prelim-top-3} we compare ours with the transfer performance and training time using top-3 sources chosen by the above methods (See Appendix I for more top-k). Among others, {\it Baseline-s} and {\it Greedy DFS} serves as the strong candidates but the training time of {\it Greedy DFS} is very high. Even so, \mym achieves the best performance using the top-3 sources and uniquely exhibits a monotonically increasing slope in all the  transfer tasks. 

\noindent{\bf Balancing Source Corpora}
In the experiements, our focus is to understand the values of the sources. For some datasets, the sizes of source corpora are very different. For example, in UD Treebank, the number of instances in Czech, and Turkish is 69k, 3.5k, respectively. 
Since data-size is an obvious factor, we conduct experiments on balanced data to reduce the influence of data-size in the analysis. 
% As our goal is to understand the value of each language and domain,
We sub-sample the source corpora to ensure their sizes are similar. Specifically, for the cross-domain NLI task, we sample 20k instances for each source. For others, we sub-sample each source such that the size of the corpus is the same as the smallest one in the dataset. However, our approach can handle both balanced or unbalanced data and the source values are similar in conclusions (e.g., see Fig \ref{fig:how-good-3datasets}).

\noindent{\bf Result:} We first compare these methods by selecting top-$k$ sources ranked by each of the approach and reporting the corresponding transfer performance.   With $k=3$,  we plot the corresponding transfer results and the running time for valuation in  Fig \ref{fig:prelim-top-3}.  As mentioned in Sec \ref{sec:intro}, the relatively strong {\it Baseline-s} can select the best performing top-1 source but with top-2 and top-3 sources, the performances drop on cross-domain sentiment analysis and cross-lingual POS tagging (See Fig \ref{fig:prelim-top-3}(c) and \ref{fig:prelim-top-3}(a)) while our approach shows a consistent gain in all of the these tasks and with top-3 sources it achieves the best performances. Appendix \ref{app:I} plots the results with higher $k$.
% excluding {\it Greedy DFS} for its long runing time. 
% Next, in Sec \ref{sec:Evaluating Source Tasks Selection}, and \ref{sec:direct-evaluation-of-source-values}, with the same \#sources selected as potential by thresholding (Sec \ref{sec:source-selection}), we further compare with other  baselines.
% {\it Baseline-s}.     

\begin{table}[t!]
	\centering
% \vspace{-3pt}
\resizebox{\linewidth}{!}{ 

	\begin{tabular}{|@{}c@{ }|c@{ }||c@{ }|c@{ }|c@{}|c@{}|c@{}|c@{}|}
		\hline
		{\bf Lang} & {\bf en} & {\bf All Source} & {\bf  Baseline-r}   & {\bf Baseline-s} & {\bf\mym}\\
% 		 &  &  &  & {\bf Shap}  &  \\
		\hline
		%=====
            en & - & {82.71} & 86.32  & 86.39 &
            {\bf 88.55$^{*\$\dag}$}
            \\
            % \hline
            % \multicolumn{7}{l}{Target Languages} \\ 
            % \hline
            % Norweian
            % no & - & 90.06 & 90.06 &
            % 90.06  & 90.06
            % \\
            % Swedish
            % sv &  83.6 & 93.26 & 
            % % 93.31 & 
            % 93.26 & 93.26
            % &
            % 93.26
            % \\
            % French
            fr & - & 94.60 & 94.63 & {\bf 94.83} &
             94.79 
          \\
        %   Portuguese
        %     pt & 82.1 & 94.33 & 94.33 &
        %   94.33 & 94.33
            % \\
            % danish
            da & 88.3 & 88.94 & 89.30  & 89.23 &
             {\bf 89.47$^{*}$}
            \\
            % Spanish
            es & 85.2 & 93.15 & 93.00  & 93.04 &
            { \bf 93.21$^{\$}$}
            \\
            % italian
            it & 84.7 & 96.58 & 96.43 & {\bf 96.71} &
            % {\bf 
            96.67
            % $^{\$}$} 
            \\
            % croatian
        %     hr & - & 96.60 & 96.60&
        %   96.60 & 96.60
        %     \\
            % catalan
            ca &  - & 91.54 & 91.64  & 90.78 &
            {\bf 92.08$^{*\$ \dag}$}
            \\
            % polish
            % pl & 86.9 & 91.61 & 
            % % 91.48 &
            % 91.61 &
            % 91.61 &
            % 91.61
            % {\bf 91.63$^\$$}
            
            % for this result 1500, 3000 num_smaples used in t-test
            % \\
            % Slovenian
            sl & 84.2 & 93.28 & 93.50 & 92.89 &
            {\bf 93.52$^{*\dag}$} 
            \\
            % dutch
            nl & 75.9 & 90.10 & 90.19 & 90.14 & 
            {\bf 90.26} 
            \\
            % Bulgarian
            % bg & 87.4 & 92.93 & 92.93 & 
            %  92.93 & 92.93
            % \\
            % russian
            ru & - & 92.98 & 92.91
             & 92.71 & {\bf 93.13$^{*\$\dag}$}
            \\
            % German
            de & 89.8 & 90.79 & { 91.07}  & {\bf \bf 91.44} & 
            91.06
            % both better than ALL sources
            \\
            % Hebrew
            he & - & 76.67 & 75.75 
             & 75.43 & {\bf 76.73$^{\$\dag}$}
            \\
            % czech
            cs & - & 93.89 & 93.04 & 93.94 & 
            {\bf 94.81$^{*\$\dag}$} 
            \\
            % romanian
            % ro &  84.7 & 89.97 & 89.97 &
            %  89.97 & 89.97
            % \\
            % slovak
            sk &  83.6 & 95.68 & 95.62  & 95.53 &
            {\bf 95.81$^{\dag}$}
            \\
            % Serbian
            sr &  - & 97.55 & 97.47  & 97.43 &
            {\bf 97.58$^{\dag}$}
            \\
            % indonesian
            id &  - & 84.10 & 85.23 & 85.50  & 
            {\bf 85.97$^{*\$}$}
            \\
            % finnish
            fi &   - & {\bf  87.13} & 86.89  & 86.86 &
            87.05
            \\
            % chinese
            % zh &  - &  71.31 & 71.31 &
            % 71.31 & 71.31
            % \\
            % ar & - & {80.07} & 80.07 & 
            % 80.07 & 80.07 
            % \\
            % Korean
            ko &  - & 63.59 & {\bf 64.27} & 63.77  &
             64.19
            \\
            % hindi
            hi &  - & 81.49 & 80.27  & 79.94 &
             {\bf 82.41$^{*\$\dag}$}
            \\
            % Japanese
            ja &  - &  66.86 & 65.99& {67.71}  & 
            {\bf 67.81$^{*\$}$} 
            \\
            % turkish
            %  tr &  - &  78.43 & 78.43 & 
            %  78.43 & 78.43 
            % \\
            % eu &  - &  81.18 & 81.18 &
            % 81.18 & 81.18 
            % \\
            fa & 72.8 & 81.03 & 80.69 
            & {\bf 82.37} &
            % {\bf
            81.79
            % $^{*\$}$} 
            \\
            \hline
            {\bf Average} & - & 82.98 & 83.05 &  83.15 & {\bf 83.66 }
            \\
		%===========
		\hline
	\end{tabular}
}
	\caption{
	\label{tab:lang-udpos}
{ Performance on universal POS tagging when using each of language as the target language and the rest as source languages .  '*', `\$', `$\dag$' denote \mym model is statistically significantly outperforms
{\it All Sources}, {\it Baseline-r} and {\it Baseline-s } respectively using paired bootstrap test with p $\leq$ 0.05. ``en'' refers to the only source (``en'') results in \citet{mbert}. %Same accuracy for all models indicates all source tasks are selected (i.e., $Z_k = D$). 
}
		}
\vspace{-10pt}
\end{table}

\begin{table}[t]

\centering
% \begin{minipage}{0.5\textwidth}
% \scriptsize

\resizebox{\linewidth}{!}{ 
 \begin{tabular}{l| l| c| c| c| c| c|c } 
 \toprule
% \hline

{\bf Model }   &  {\bf  WSJ} & {\bf EM } & {\bf  N } &  {\bf A}  &  {\bf R} &  {\bf WB} & {\bf Avg}\\
% \midrule
\hline
MMD & 96.12   & 96.23 & 96.40 & 95.75 & 95.51 & 96.95 & 96.16 \\
RENYI & 96.35  & 96.31 & 96.62 & 95.52 & 95.97 & 96.75 & 96.25\\
\midrule

All Sources & 95.95 &  95.39 & 96.94 & 95.15 & 96.08  &  97.10 & 96.10 
\\
Baseline-r& 95.98 & 93.41 & 93.78 & 93.14 & 95.25  & 97.10  & 94.78
\\
\mym & {\bf 96.14$^{*\$}$} & {\bf 95.47$^{\$}$} & {\bf 97.02$^{\$}$} & {\bf 95.30$^{*\$}$} & {\bf 96.17$^{\$}$} & 97.10 & {\bf 96.20} 
\\
 \bottomrule
% \hline

 \end{tabular}
 }
 \caption{ POS tagging results (\% accuracy) on SANCL 2012 Shared Task. '*'  and `\$' denote the model using \mym statistically significantly outperforms
{\it All Sources} and {\it Baseline-r } respectively using paired bootstrap test with p $\leq$ 0.05.
 MMD,  and RENYI refer to \citet{reinforced} which use auxiliary unlabelled data in the target domain and focus on instance selection. {\it Baseline-s} has exactly same results as \mym. 
 }
 \label{tab:dom-pos}
%  \end{minipage}
% \vspace{-18pt}
\end{table}

\setlength{\tabcolsep}{3pt}
\begin{table}[h]

\centering
\resizebox{\linewidth}{!}{ 

  \begin{tabular}{ l | c| c| c| l |  l| c| c| c| l } 
 \toprule
{\bf Model }   &    {\bf bg } & {\bf ru } & {\bf tr } & {\bf ar } & {\bf vi } &  {\bf hi } & {\bf sw } & {\bf ur } & {\bf Avg}\\
\midrule
 XLM-MLM  & 74.0 & 73.1 & 67.8 & 68.5 & 71.2 & 65.7 & 64.6  & 63.4 &  68.54 
\\
mBERT(en) & 68.9 &  69.0 & 61.6 & 64.9 & 69.5 & 60.0 &   50.4 & 58.0 & 62.79
\\
\midrule
All Sources &  74.03 &  73.59 & 65.21 & 68.94  & 74.39  & 67.31 & 52.67  & 64.37 & 67.56
\\
Baseline-r&   74.69 & {\bf 74.53} & {\bf 65.85} & 68.68 & 75.03  & 66.69 & {\bf 52.97} & 63.69 &  67.77
\\
Baseline-s &    73.23 &   73.73  & {65.67} & {68.36} & {74.11} &  {67.07} & { 52.59} & {63.31} & 67.26
\\
Ours &  {\bf 74.95} &   73.85  & { 65.63} & {\bf 69.24} & {\bf 75.71}  & {\bf 67.78} & { 52.73} & {\bf 64.67} & {\bf 68.07}
\\
 \bottomrule
 \end{tabular}
 }
\caption{{  XNLI results. As a reference, we include two results from the recently published papers mBERT~\cite{mbert} and ``XLM-MLM''~\citet{XLM}. mBERT is trained on ``en'' only and ``XLM-MLM'' is applicable to XNLI languages only.   
 }}
 \label{tab:lang-xnli}
\vspace{-10pt}
\end{table}

\setlength{\tabcolsep}{3pt}
\begin{table}[h]
\label{tab:UPOS}
\centering
\resizebox{\linewidth}{!}{ 
\begin{tabular}{l| l| c| c| c|  c | r  } 
 \toprule
{\bf Model }   &  {\bf books} & {\bf kitchen} & {\bf dvd} &  {\bf baby} &  {\bf   MR} & {\bf Avg}
\\
 \midrule
\citet{ijcaibooks} &  87.3 & 88.3 &  88.8   & 90.3  & 76.3  & 86.2
\\
\midrule
% \hline 
All Sources & {\bf 87.3} & 90.3 & 88.3 & 92.3 & 79.3 & 87.5
\\
Baseline-r & 87.0 & 90.5 & 87.3 & 91.8 & 78.8 & 87.1
\\
Baseline-s & 86.8 & 89.8 & 87.0  & {\bf 92.5} &  77.5  &  86.7
\\
\mym & {\bf 87.3} & {\bf 90.8} & {\bf 88.8} & {\bf 92.5}  & {\bf 79.5} & {\bf 87.8}
\\
 \bottomrule
 \end{tabular}
 }
 \caption{ Cross-domain transfer results on multi-domain sentiment analysis task. \citet{ijcaibooks} use unlabelled data from the target domain. 
 }
 \label{tab:domain-mtl-senti-main}
%  \end{minipage}
\vspace{-0.2cm}
\end{table}

\setlength{\tabcolsep}{5pt}
\begin{table}[h]
% \begin{minipage}{0.5\textwidth}
\scriptsize
\centering
\resizebox{\linewidth}{!}{ 
 \begin{tabular}{l| l| c| c| c| c} 
 \toprule
% \hline
{\bf Model }   &   {\bf  SNLI} & {\bf QQP } & {\bf  QNLI } & {\bf MNLI-mm} & {\bf Avg}\\
% \midrule
\hline
\citet{bert-domain} & 88.30 &73.90 & 59.10 & - & 76.23
\\
% \midrule
\hline
All Sources & 88.69  & 72.96 & 50.65 & 89.47 & 75.45
\\
Baseline-r& 88.11 & 72.71 & 50.53 & 89.18 & 75.13
\\
Baseline-s& {\bf 88.72} & {\bf 73.47} & 50.98 & {\bf 89.69} & 75.72
\\
\mym & {\bf 88.72} & {\bf 73.47} & {\bf 54.75} & {\bf 89.69} & {\bf 76.66}
\\
 \bottomrule
% \hline
 \end{tabular}
 }
 \caption{{ Zero-shot results on modified GLUE. \citet{bert-domain} selects instances from one source domain at once while we select a subset of source corpora.  
 }}
 \label{tab:dom-nli}
%  \end{minipage}
 \vspace{-10pt}
\end{table}

% \subsection{Evaluating Source Corpora Selection }
% \label{sec:Evaluating Source Tasks Selection}
Next, as in Sec \ref{sec:source-selection}, we tune a threshold $\theta$ and  either select all the sources as useful or a smaller subset of $m$ number of sources (i.e., $m<|\mathcal{D}|$) whose \mym values are higher than  $\theta$. In the followings, we compare the model performances of these $m$ sources selected by \mym with the same top-$m$ sources ranked by the aforementioned baseline methods. 
Being relatively weak or slow, we do not further report performances for {\it LOO}, {\it Lang-Dist},  and {\it Greedy DFS}. Rather we consider another strong baseline {\it All Sources} that uses all the source corpora $\mathcal{D}$.
% Note that {\it All Sources} is a strong baseline as it trained on more source-corpus instances than \mym in general. 
This is a strong baseline as it is trained on more source-corpus instances
%  than \mym 
 in general. 
%  Note that \mym performance is same as {\it All Sources} when selecting all sources $\mathcal{D}$ as useful.} 
% \paragraph{Cross-Lingual POS Tagging}

\noindent{\bf Cross-Lingual POS Tagging} 
% Table \ref{}
\label{sec:Cross-lingual Universal POS Tagging}
We evaluate the source selection results on zero-shot cross-lingual POS tagging in Table \ref{tab:lang-udpos}. Among the 31 target languages, in 21 of  them, \mym selects a small subset of source corpora. From the Table, overall, \mym selects source corpora with high usefulness for training the model, and except for few cases the model constantly outperforms all the baselines by more than
$0.5\%$ in avg token accuracy.
In 13 of them, it is statistically significant by a paired bootstrap test. 
The gap is especially high for English,  Czech, and Hindi. 
% When using the same size of data, \mym is significantly better than  {\it Baseline-s} in 13 languages.
These results demonstrate that \mym is capable in both quantifying the source  values and also in source selection. We report the full results on the dev and test set of all target languages in Appendix  \ref{app:M},  \ref{app:N} respectively. For each row in Table \ref{tab:lang-udpos},  the number of selected sources are reported in Appendix  \ref{app:S}. 
%As for the random selection baseline  {\it Baseline-s},
%although {\it Baseline-s} manages to achieve high accuracy than {\it All Sources } for some languages, the predictions of our framework 
%consistently surpasses it as well on all the same 20 languages except for 2 where {\it Baseline-s} achieves slightly higher accuracy than ours. Nonetheless, as the improvement in POS tagging tends to be numerically little, we perform an additional significance test.    as in Table \ref{tab:lang-udpos}, among the 20 languages where \mym outperforms others, on 8 of them, it statistically significantly better than both baselines and on 5 of them, it is statistically significantly better than one of the two baselines. 
\noindent{\bf Cross-Domain POS Tagging }
Table \ref{tab:dom-pos} presents the POS tagging results in  zero-shot domain transfer on SANCL 2012 shared task. In 5 out of 6 targets, \mym outperforms all baselines except {\it Baseline-s}. For each target domain with only 5 sources, {\it Baseline-s} source values match with ours in general. 
% NOTE THAT NEED TO MENTION THIS IN CAMERA READY
% In one CASE their order is slightly different top-
 However,
\mym significantly outperforms {\it Baseline-r} on all 5 cases and   {\it All-Sources} twice.  %In statistical significant test, on all of them, it is better than {\it Baseline-s} and on 2 of them it is better than both.  
It even outperforms MMD, and RENYI~\cite{reinforced} on Newsgroups (N), Reviews (R), and Weblogs (WB) despite they
 select source data at instance level and use additional resources.  
% it achieves the state-of-the-art performance. Additionally, although we are focused on domain selection, it achieves competitive performance with instance selection approach \cite{reinforced}. 
\begin{figure}[t]
\vspace{-0.15cm}
% \hspace{-0.1cm}
\centering
\includegraphics[width=0.95\linewidth]{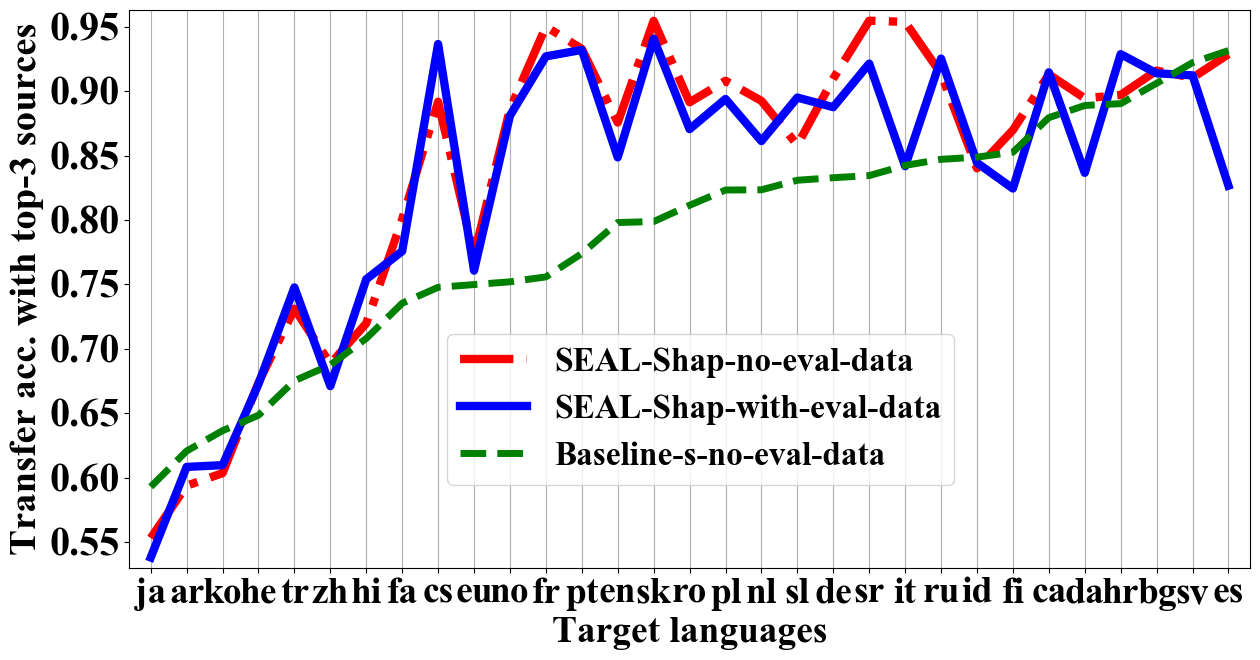}
\vspace{-0.3cm}
\caption{
% \vspace{1.5cm}
{\small {Cross-lingual POS tagging accuracies on different target languages using top-3 sources ranked by \mym. 
The ranker (red) selects similar sources as using \mym with annotated target data (blue).
Ranker trained to predict \mym values (red) performs better than baseline (green)~\cite{neubig-choosing}.}}
}
% \vspace{-0.4cm}
  \label{fig:linetal-ranker}
\end{figure}
% \paragraph{Cross-Domain POS Tagging}
% \paragraph{Cross-Lingual NLI}

\noindent{\bf Cross-Lingual NLI}
% Along with the cross-lingual POS tagging, we also consider the  cross-lingual NLI task on XNLI dataset.
% Results show that in 7 out of 15 target languages, all source languages are positively contributed to  the transfer performance estimated by \mym. Therefore, \mym is the same as \textit{All Sources} in those cases. This is possible as NLI task is a semantic-oriented task and the variance between languages is small.
% We ob, having more source data is in general helpful. 
In Table \ref{tab:lang-xnli}, we show the XNLI results in 8 target languages  where \mym selects a small subset of source corpora. Among them, in 3 languages, {\it Baseline-r} marginally  surpasses ours. However, in 5 other languages \mym outperforms all the baselines with clear margin specially on Bulgarian, Vietnamese  with about 1\% better accuracy (full results in Appendix \ref{app:E}). 
% showing the \mym is still able to identify valuable source corpora even in this extreme case. 
% Full results on all target languages are in Appendix E        
% \paragraph{Cross-Domain NLI }

\noindent{\bf Cross-Domain NLI}
 Next, we evaluate \mym on the modified GLUE dataset in Table \ref{tab:dom-nli}. \mym outperforms {\it Baseline-s} once and other baselines in all cases.
 %and competitive with \citet{bert-domain} even though allows to select source data at the instance level while \mym only selects data at the corpus level. 
 %Selecting the source domains randomly ({\it Baseline-s} significantly falls short than others. On {\color{blue} N} target domains, the BERT classifier trained with source domains selected by our approach outperforms both baselines.
 Its highest performance improvement is gained on QNLI, where it outperforms others by $4\%$.
%  An interesting outcome is that although both QQP and QNLI consist of question based sentences, for target QNLI, \mym selects only QQP as a potential source, but for target QQP, it selects all other domains except QNLI. The single source transfer performances reported in \cite{bert-domain} also validate this result. For target QNLI, QPP has the highest single source result but for target QQP, QNLI has the lowest score.
%we observe that out of 15 target languages, in 9 languages, our approach finds that all the source languages are actually potential. Only in  6 languages, our framework selects a smaller subset. In Table  \ref{tab:lang-xnli}, we present the results on these  6 languages (full results are in Appendix).  As can be seen in the Table, although it outperforms the baseline {\it All Sources}, the improvement is relatively small.
%Note that, due to the extremely high data volume of XNLI dataset, we only consider the {\it All Sources} as baseline.  
% \paragraph{Cross-Domain Sentiment Analysis}

\noindent{\bf Cross-Domain Sentiment Analysis}
Among the 13 target domains in the multi-domain sentiment analysis dataset, in 5 domains \mym selects a small subset (full results in Appendix \ref{app:O}). As in Table \ref{tab:domain-mtl-senti-main}), with a large margin, \mym achieves higher accuracy than all other baselines and, in 4 cases, it is even better than \citet{ijcaibooks} that uses unlabeled target data. 
% In some case, \mym selects all sources as all of them are contributed positively to the transfer performance. In such a case, \mym reports same performance as \textit{All Sources}. 

%and  3 other domains, it matches the performance of an existing best.  

 \begin{figure}[t]
\vspace{-0.38cm}
% \hspace{0.4cm}
\centering
\includegraphics[width=0.82\linewidth, height=0.85\linewidth]{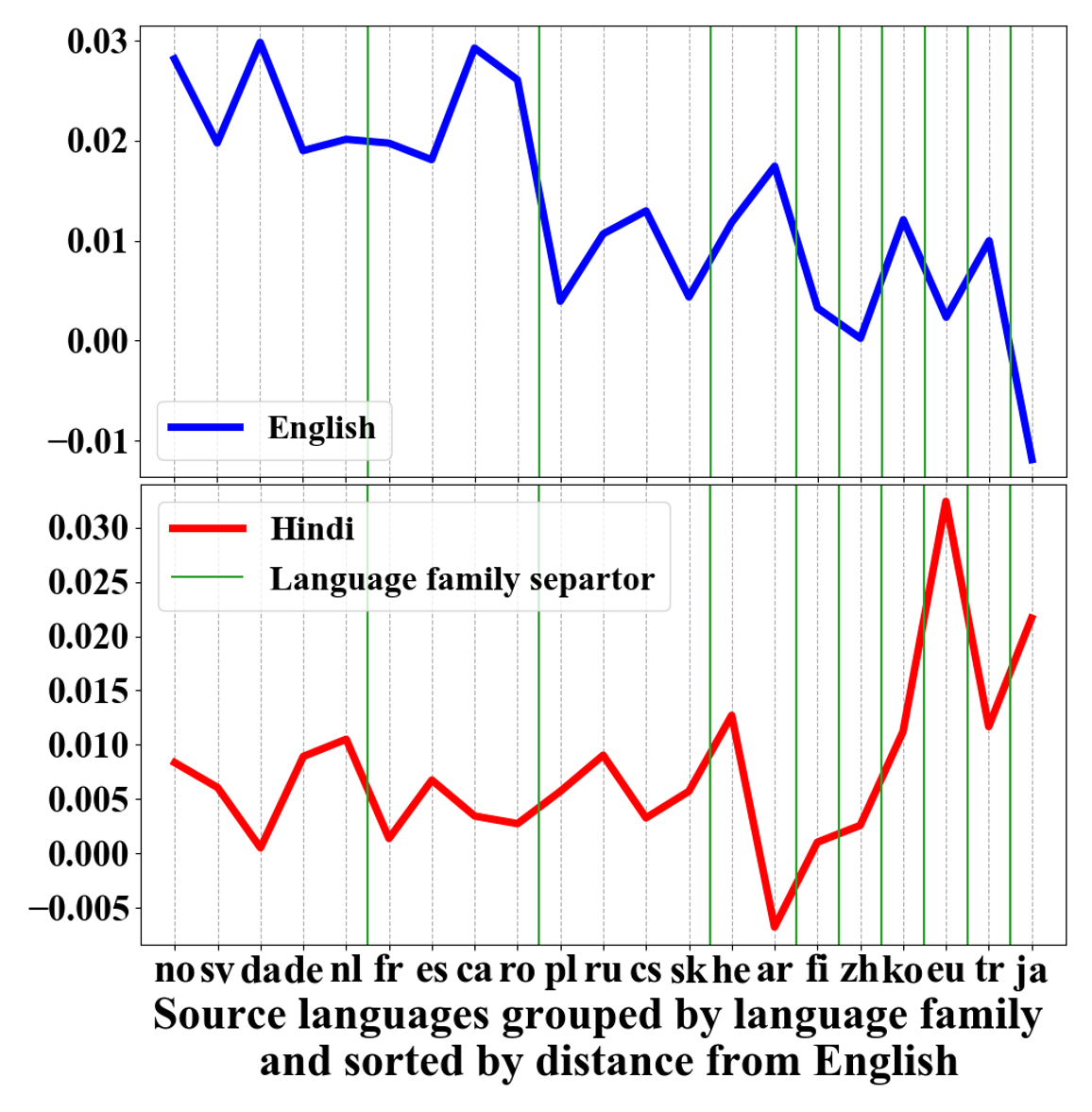}
\vspace{-0.3cm}
\caption{
{\small {Cross-lingual POS tagging \mym values, referring to the relative contribution of the source languages.}}
\vspace{-0.40cm}
}
  \label{fig:interpret}
\end{figure}

\begin{figure*}[h]
\setlength{\abovecaptionskip}{-4pt}
\vspace{-0.4cm}
  \centering
  \resizebox{\linewidth}{!}{ 
    \begin{tabular}{c@{}c@{}}
    \hspace{-0.651cm}
% \vspace{-1.5cm}
  \subfigure[XNLI, target: 'es',  R $<$10$\%$]{\
        \includegraphics[height=0.26\linewidth]{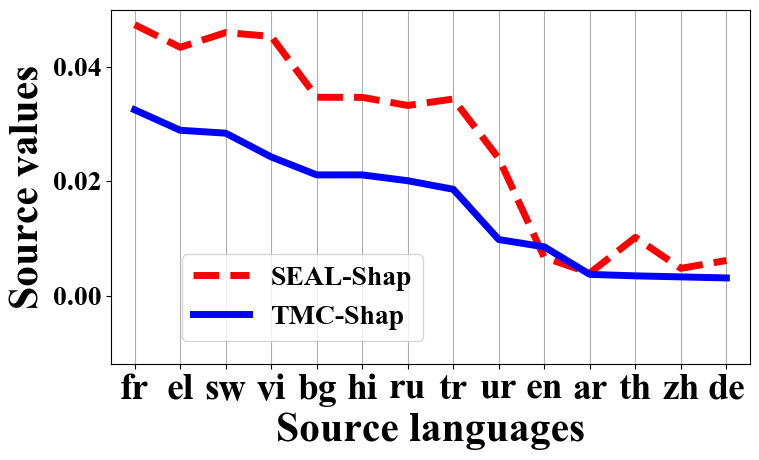}\label{fig:skim-comp-imdb}
        } &
        % \hspace{-0.57cm}
        \subfigure[mGLUE, target: MNLI-mm, R=10-20$\%$]{\includegraphics[height=0.26\linewidth]{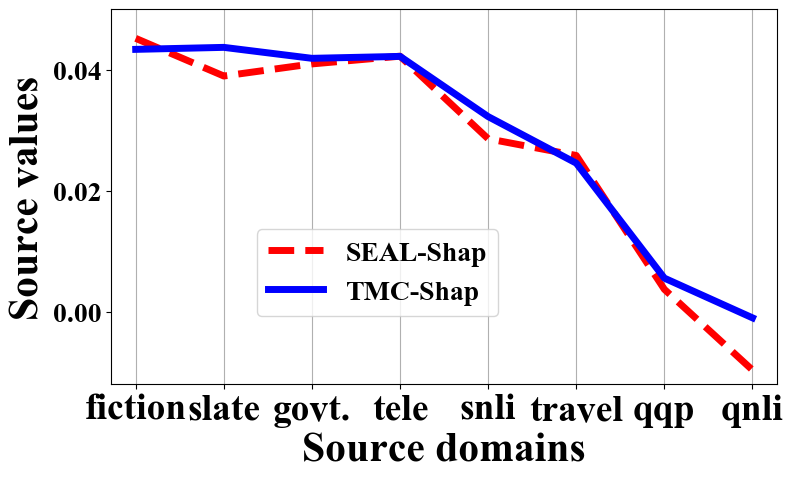} \label{fig:skim-comp-agnews}} 
%         &
% \hspace{0.02cm}
% \hspace{1pt}
  \subfigure[SANCL'12, target: wsj, R $\sim$50\%]{\includegraphics[height=0.26\textwidth]{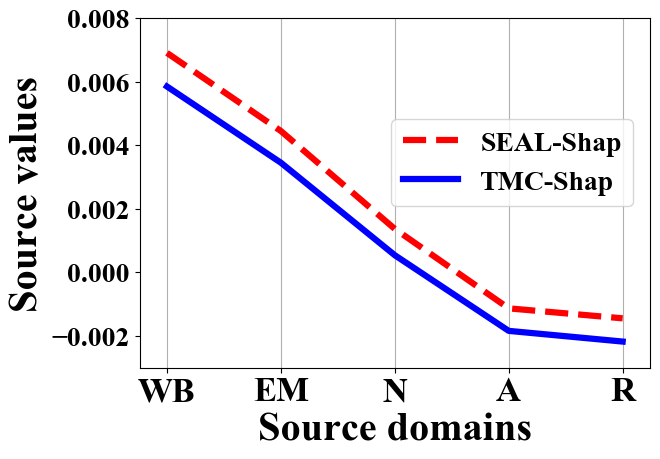}\label{fig:imdb_latm_sel}}
    \end{tabular}
    }
  \caption{ 
  {{ Source values by TMC-Shap and ours. TMC-Shap uses unbalanced full source corpora whereas \mym that achieves  similar source values uses balanced and sampled source corpora. Even with a small sample rate (R), source order is  almost same. Higher sampling rate typically refers to better approximation but leads to expensive runtime. In general, for a reasonably large corpus, 20-30\% samples (>few thousands) are found sufficient to achieve reasonable approximation.  
}
}   
  } 
  \label{fig:how-good-3datasets}
  \vspace{-5pt}
\end{figure*}

Our experimental evidences show that \mym is an effective tool in choosing useful transfer sources and can achieve higher transfer performances than other source valuation approaches.

\subsection{Results without an Evaluation Corpus}
\label{sec:direct-evaluation-of-source-values}
We evaluate the effectiveness of \mym to build a straightforward ranker  that directly computes the source values without any evaluation target corpus (see Sec \ref{sec:direct-source-value}). 
We use the ranker in \citet{neubig-choosing} as the underlying ranking model. First, we show that the source values evaluated by the ranker is as good as \mym that uses its annotated target dataset. We compare the transfer performances of the top-$k$ sources based on the source values computed with and without the evaluation corpus. Then, we show that the ranker trained with \mym is more effective than training it with the existing single source based {\it Baseline-s}.

In  cross-lingual POS tagging on UD Treebank, for each of the 31 target languages, we set aside that language and consider the remaining 30 languages as the training corpora. We then train the ranker as described in Sec \ref{sec:direct-source-value} and compute the source values using it.  As for reference, we pass the evaluation target dataset and the 30 source languages to \mym to compute their values on the evaluation  dataset. With $k=3$, we compare the transfer results of the top-$k$ sources of these two methods
in Fig \ref{fig:linetal-ranker}. 
We also plot the results of the baseline ranker \cite{neubig-choosing} that is trained with {\it Baseline-s}. 
Results show that the ranker source values are similar to the sources values estimated by \mym  with an annotated evaluation dataset
and also it outperforms the baseline.

% Fig \ref{fig:linetal-ranker} shows that the ranker source values are similar to \mym  it outperforms the results of the baseline ranker, trained with {\it Baseline-s} \cite{neubig-choosing}.

\subsection{Interpret Source Value by \mym}
\label{sec:interpretablity}

% {\color{blue}
% Extend this section a bit, as this would be the most interesting section for this paper. You can move some discussion currently you have in 3.3 to here. You can orgainze this section as what you have in your current section 4 (ablation QA). 
% }

In this Section, we show that SEAL-Shap values provide a means to understand the usefulness of the transfer sources in cross-lingual and cross-domain transfer.
We first analyze 
%To visualize this, we take an example of 
cross-lingual POS tagging.
Following \citet{on-difficulties}, we consider using language family and word-order distance as a reference distance metric. We anticipate that languages in the same language family  with smaller word-order distance from the target language are more valuable in multi-lingual transfer.
%, while languages in the same language family as the target language are more valuable.
%High word order distance, ideally, refers to a very distant language. Therefore, it can be anticipated that given a particular problem like POS tagging, source languages of the similar language family and closer in word order distance might be more informative for the corresponding target task. Now, in 
We plot \mym  of source languages evaluated on two target languages English  (``en'') and Hindi (``hi'') in Fig \ref{fig:interpret}. In the x-axis, a common set of twenty different source languages are grouped into ten different language families and sorted based on the word order distance from English. As the figure illustrates,  
%Shapley values for English target language are computed high for 
Germanic and Romance languages have higher Shapley values when using English as the target language. The value gradually decreases for language of other families when the  word order distance increase. As for the target language Hindi, the trend is opposite, in general. 
%One outlier of this trend is Arabic (``ar''). To investigate even more, we perform an example case-study on why Shapley values of language Arabic (``ar'') might be higher than Hindi and found that as per the World Atlas of Language Structures (WALS) \cite{wals}, in Arabic, verb  precedes object whereas in Hindi it is opposite. 
% Analogously,  in cross-domain NLI, we find that correlation between QNLI, and QQP is high  whereas between MNLI-mm and  QQP, it is lower (see Appendix Q).
 \begin{figure}[h]
\vspace{-0.2cm}
% \hspace{-0.535cm}
\centering
\includegraphics[width=\linewidth]{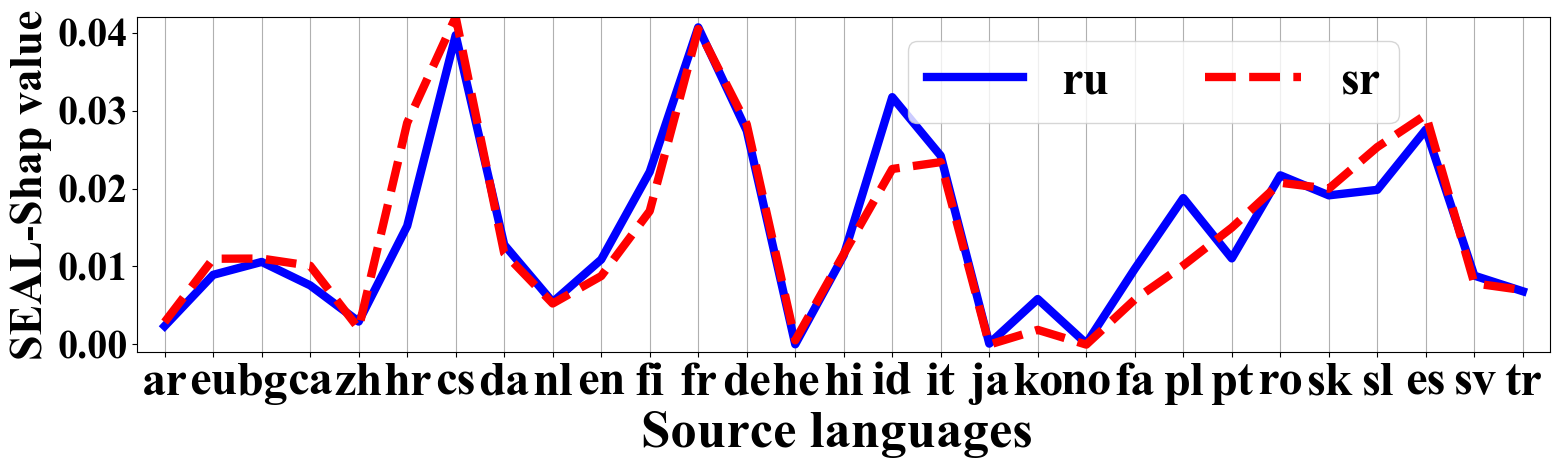}
\vspace{-0.8cm}
\caption{
% \vspace{1.5cm}
{\small {Similar \mym value curves for two closely related  target languages in cross-lingual POS tagging.}}
% \vspace{-0.2cm}
}
  \label{fig:two-sim-ru-sr}
\end{figure}
 \begin{figure}[h]
\vspace{-0.5cm}
\centering
\includegraphics[width=0.75\linewidth,height=0.4\linewidth]{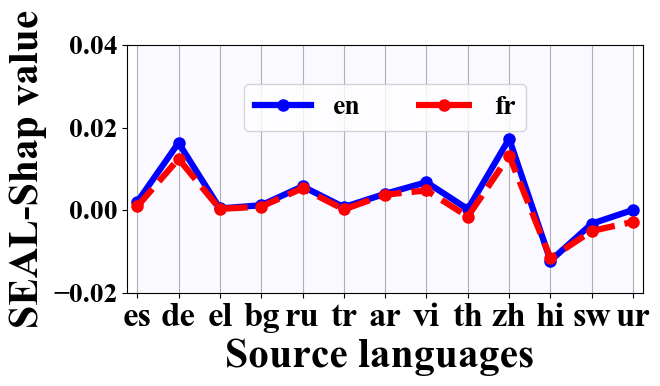}
\vspace{-0.3cm}
\caption{
% \vspace{1.5cm}
{\small {Similar \mym value curves for two closely related  XNLI targets ``en'' and ``fr''. In XNLI, the source corpora are prepared by machine translating from ``en''. This data processing may affect the source values. Translation into ``zh'' being relatively better, although different from both targets, its source values are higher than others.}}
\vspace{-0.3cm}
}
  \label{fig:two-sim-en-fr}
\end{figure}

Analogously,  in cross-domain NLI, we find that correlation between QNLI, and QQP is high  whereas between MNLI-mm and  QQP, it is lower (see Appendix \ref{app:Q}).
% \paragraph{\mym on Similar Targets}

\noindent{\bf \mym on Similar Targets}
Intuitively, if two target corpora are similar, the corresponding Shapley values of the source corpora when transferring to these two targets should be similar as well. To verify, in Fig \ref{fig:two-sim-ru-sr}, we plot the Shapley values of  twenty nine source languages for targets Russian and Serbian on cross-lingual POS tagging. Also we plot the source values when transferring a NLI model to English and French in Fig \ref{fig:two-sim-en-fr}. We observe that the corresponding curves are almost identical, and \mym in fact selects the same set of source corpora as potential. These results suggest that if there is no sufficient data in the target corpus, it is also possible to use a neighboring corpus as a proxy to compute \mym values.

 \noindent{\bf Source Values Influenced by Data Processing} Typically, the sources with least or negative source values are from the domains/languages that are different from the targets (e.g., Fig \ref{fig:interpret}). However, in some cases, source usefulness (i.e., values) is affected by the data preparing process. For example, in XLNI, the source corpora are prepared by machine translation from ``en''~\cite{xnli} and the quality of this translation into ``zh'' is better in compare to other languages, in general. Consequently, in Fig \ref{fig:two-sim-en-fr}, ``zh'' has higher source value for both targets ``en'' and ``fr''.

% \section{Analysis}

\setlength{\tabcolsep}{1pt}
\begin{table}[t!]
\scriptsize
\centering
%  \vspace{-15pt}
\resizebox{\linewidth}{!}{ 
 \begin{tabular}{l| l| c| c| c| c| c} 
 \toprule

\multirow{2}{*}{\bf Prob.} & \multirow{2}{*}{\bf Transfer} & \multirow{2}{*}{\bf Target} & \multirow{2}{*}{\bf \#Targets} & \multirow{2}{*}{\bf \#Samples} & \multirow{2}{*}{\bf  Caching } &  {\bf Time } 
\\
& &  &   &  & &  {\bf (hours)}  
\\
\midrule
% \hline
 \multirow{4}{*}{\bf NLI} & \multirow{4}{*}{\bf Domain} & \multirow{4}{*}{\bf MNLI-mm} & 1 & \xmark & \xmark & 300$^*$
\\
% \hline
 & &  &  1 & \xmark & \cmark & 101
\\
 & &  & 1 & 20k & \cmark & 18
\\
 & &  & 3 & 20k & \cmark & 5
 \\
 \midrule
 
  \multirow{4}{*}{\bf POS} & \multirow{4}{*}{\bf Language} & \multirow{4}{*}{\bf Arabic (ar)} & 1 & \xmark & \xmark & 210$^*$
\\
% \hline
 & &  &  1 & \xmark & \cmark & 180$^*$
\\
 & &  & 1 & 3.3k & \cmark & 25
\\
 & &  & 31 & 3.3k & \cmark & 3.5
 \\
 \bottomrule
 \end{tabular}
 }
 \caption{{ Running time for computing approximate Shapley value. The marker $^*$represents the time is estimated by extrapolation. \#Targets indicates number of target corpus evaluated simultaneously. \#Samples is the number of samples used to train model for computing marginal contribution. TMC-Shap is equivalent to disable all the techniques (the first row of each block). 
% For each Problem, the first row refers to  the TMC-Shap. 
 }}
 \label{tab: time-save}

\end{table}

\subsection{Analysis and Ablation Study}
\label{sec:ablation}

Finally, we analyze the proposed Algorithm \ref{algo-1} for  computing Shapley value approximately. %In this section, we discuss several properties of our approach by answering a few question below.

% \paragraph{How good is the approximation?}
\noindent{\bf How good is the approximation?}
In Fig \ref{fig:how-good-3datasets}, we compare \mym with TMC-Shap~\cite{james-data-shapley} on three datasets (details  in Appendix \ref{app:F}).  
%on three example datasets with three different sample size we compute the approximated data Shapley and compare with the existing TMC-Shap approach \cite{james-data-shapley} (details in Appendix). 
Overall, the Shapley values obtained by \mym and TMC-Shap are highly correlated and their relative orders are matched, while \mym is much more efficient. Note that, the rankings themselves being same/similar, the model performances using the same/similar top-$k$ sources are same/similar, too; therefore, we do not list their transfer performances furthermore. 

% For example, for target (R)eviews in Table 2, the top-4 rankings of TMC-Shap and SEAL-Shap are [A, WB, N, E], and [A, N, WB, E] and performances using upto top-4 sources are [0.957, 0.9577, 0.9584, 0.961], and [0.957, 0.9579, 0.9584, 0.961], respectively.

%obtain similar Shapley value for each source and the 

%matches with the relative order of the source tasks very well. In fact, for XNLI while we consider only 50k samples (rate $<10\%)$ from each source languages, but the order mismatches on only one source language ``th''. 

% \paragraph{Ablation Study} 

\noindent{\bf Ablation Study:} 
% To answer these,  
 We examine the effectiveness of each proposed components in \mym. Results are shown in Table \ref{tab: time-save} and details are in Appendix \ref{app:F}-\ref{app:H}. Results show that without the proposed approximation, TMC-Shap is computational costly and is impractical to use to analyze the value of source corpus in the NLP transfer setting. All the proposed components contribute to significantly speed-up the computations. 
 %Shapley value computation time needed with/out different factors (e.g., multi-target evaluation, caching etc.,) (details in Appendix). In Table \ref{tab: time-save}, 
 
% the top row for each problem is equivalent to TMC-Shap and last row is our highest time-saving. As it can be seen in the Table, sampling of the source tasks leads to highest speedup. 

\begin{figure}[t!]
\vspace{-5pt}
\centering
\includegraphics[width=\linewidth]{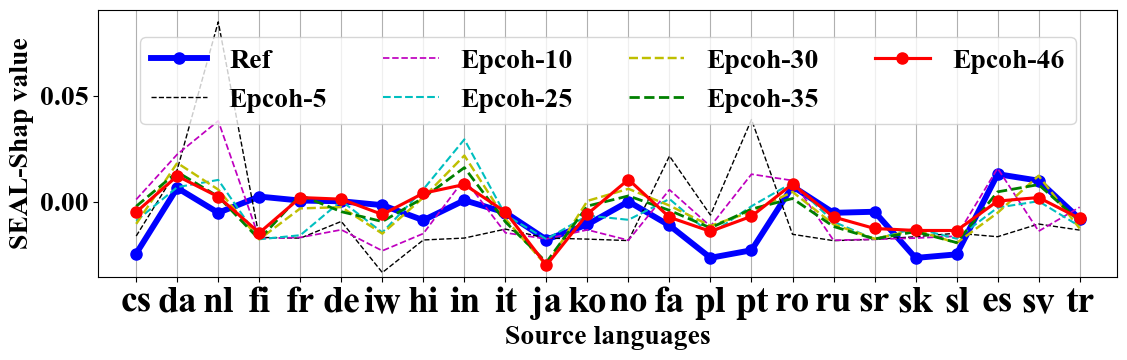}
\vspace{-20pt}
\caption{ \mym value with two (colored) seeds.} 
\vspace{-0.4cm}
  \label{fig:shap-diff-seed-main-paper}
\end{figure}

% \paragraph{Is the approximation sensitive to the order of permutations? }
\noindent{\bf Is the approximation sensitive to the order of permutations?}
%As discussed in Section \ref{sec:ftmc-shap}, while computing the Shapley value,  for  an early convergence or  to get a better estimation within given ``nepoch'', we propose to use a high initial score.  Now, the question arises that given a certain ``nepoch'', is 
As \mym is a Monte Carlo approximation, we study if \mym is sensitive to the random seed using the cross-lingual POS tagging task. To analyze, we first compute a reference Shapley values by running \mym until empirically convergence (blue line). Then, we report the Shapley value produced by another random seed. Fig \ref{fig:shap-diff-seed-main-paper} shows that with enough epochs, the values computed by different random seeds are highly correlated (more in Appendix  \ref{app:H}).

%our approximation sensitive to the or  order of permutation. To answer this,  we perform a case study with a reasonably large ``nepoch'' with two different seeds. First, we finish computation using a reference seed. Then we start with the other seed and  record the approximated Shapley value after each epoch. In Appendix, we show that gradually with the increase of epoch, both approximation converges. Hence, given the ``nepoch'' reasonably large, it not sensitive to different seeds or different order of permutations. However, when the number of source tasks are high, multiple seeds can be beneficial. 

\section{Related Work}
\label{sec:rel-work}

As discussed in Section \ref{sec:intro}, transfer learning has been extensively studied in NLP to improve model performance in low-resource domains and languages. In the litearture, various approaches have been proposed to various tasks, including text classification \cite{zhou2016transfer, kim2017cross}, natural language inference~\cite{Conneau-lample2018word,artetxe2019massively},  sequence tagging \cite{tackstrom-etal-2013-token, agic2016multilingual, kim2017cross,  ruder-plank-2017-learning}, dependency parsing \cite{guo2015cross,meng2019target}. These prior studies mostly focus on bridging the domain gap between sources and targets.
%selecting individual instance and the value of a whole source corpus is under explored specially in different source combinations. 

In different contexts, methods including influence functions and Shapley values have been applied to value the contribution of training data~\cite{koh2017understanding, lundberg2018consistent, knn-shapley-vldb}. 
Specifically, Monte Carlo approximation of Shapley values has been used in various applications~\cite{MC_ONLY_METHOD_FOR_LARGE_SCALE_SHAPLEY, knn-shapley-vldb, ghorbani2020neuron, tripathi2020feature, tang2020data,Many_shapley}. However they are either task/model specific or not scalable to NLP applications. Oppositely, \citet{shapley2020problems} discuss the problems of using Shapley value for model explanation. In contrast, we apply efficient Shapley value approximation in NLP transfer learning and analyze the source-target relationships.   
\section{Conclusion}
We propose \mym to quantify the value of the source corpora in transfer learning for NLP by  computing an approximate Shapley value for each corpus.
%build a novel transfer learning framework that considers the data Shapley value to quantify the usefulness of the source tasks. We extends an existing Shapley value approximation method to handle practical large scale NLP settings. 
We show that \mym can be used to select source corpora for transfer and provide insight on understanding the value of source corpora. 
%using the subset of source tasks chosen by our framework significantly better than using all the source tasks.  And the quantification promotes the source selection procedure interpretable.
In the future, we plan to further improve the run-time of our source valuation approach by limiting the repetition of model training.     

%explore robust representation learning and bias mitigation using the approximated data Shapley value. 

\section{Acknowledgments}
We thank the anonymous reviewers for their insightful feedback. We also thank UCLA-NLP for discussion and feedback. This work
was supported in part by NSF 1927554 and DARPA MCS program under Cooperative Agreement N66001-19-2-4032. The views expressed are those of the authors and do not reflect the official policy or position of the Department of Defense or the U.S. Government.
% \input{ethics}

% Use unnumbered third level headings for the acknowledgments. All acknowledgments, including those to funding agencies, go at the end of the paper.

\bibliography{emnlp2020,nlp}
% \bibliography{preprint}
% \bibliography{ref}
\bibliographystyle{acl_natbib}

% moved to supplementary material
% REMOVE the comments to add Appendix for Arxiv version
\cleardoublepage
\appendix
\onecolumn
\section*{Supplementary Material: Appendices}

\addcontentsline{toc}{section}{Appendices}
\renewcommand{\thesubsection}{\Alph{subsection}}

\subsection{Details of UD Treebanks}
\label{app:langstat}

We use the \href{https://github.com/flairNLP/flair/blob/master/resources/docs/TUTORIAL_6_CORPUS.md}{Flair} framework provided version of UDTreebank.

The statistics of the Universal Dependency treebanks (v2.2) is summarized in Table \ref{tab:udstat}. However, more accurate statistics can be found using the above link.

\begin{small}
	\begin{longtable}{l | l | c | c c|  c} 
    \hline
		Language & Lang. Family & Treebank & Num. of & Sent. & \#Token(w/o punct)\\
		\hline
		\multirow{3}{*}{Arabic (ar)} & \multirow{3}{*}{Afro-Asiatic} & \multirow{3}{*}{PADT} & train & 6075 & 223881(206041)\\
		& & & dev & 909 & 30239(27339)\\
		& & & test & 680 & 28264(26171)\\
		\hline
		\multirow{3}{*}{Bulgarian (bg)} & \multirow{3}{*}{IE.Slavic} & \multirow{3}{*}{BTB} & train & 8907 & 124336(106813)\\
		& & & dev & 1115 & 16089(13822)\\
		& & & test & 1116 & 15724(13456)\\
		\hline
		\multirow{3}{*}{Catalan (ca)} & \multirow{3}{*}{IE.Romance} & \multirow{3}{*}{AnCora} & train & 13123 & 417587(371981)\\
		& & & dev & 1709 & 56482(50452)\\
		& & & test & 1846 & 57902(51459)\\
		\hline
		\multirow{3}{*}{Chinese (zh)} & \multirow{3}{*}{Sino-Tibetan} & \multirow{3}{*}{GSD} & train & 3997 & 98608(84988)\\
		& & & dev & 500 & 12663(10890)\\
		& & & test & 500 & 12012(10321)\\
		\hline
		\multirow{3}{*}{Croatian (hr)} & \multirow{3}{*}{IE.Slavic} & \multirow{3}{*}{SET} & train & 6983 & 154055(135206)\\
		& & & dev & 849 & 19543(17211)\\
		& & & test & 1057 & 23446(20622)\\
		\hline
		\multirow{3}{*}{Czech (cs)} & \multirow{3}{*}{IE.Slavic} & \multirow{3}{*}{{PDT,CAC,CLTT,FicTree}} & train & 102993 & 1806230(1542805)\\
		& & & dev & 11311 & 191679(163387)\\
		& & & test & 12203 & 205597(174771)\\
		\hline
		\multirow{3}{*}{Danish (da)} & \multirow{3}{*}{IE.Germanic} & \multirow{3}{*}{DDT} & train & 4383 & 80378(69219)\\
		& & & dev & 564 & 10332(8951)\\
		& & & test & 565 & 10023(8573)\\
		\hline
		\multirow{3}{*}{Dutch (nl)} & \multirow{3}{*}{IE.Germanic} & \multirow{3}{*}{{Alpino,LassySmall}} & train & 18058 & 261180(228902)\\
		& & & dev & 1394 & 22938(19645)\\
		& & & test & 1472 & 22622(19734)\\
		\hline
		\multirow{3}{*}{English (en)} & \multirow{3}{*}{IE.Germanic} & \multirow{3}{*}{EWT} & train & 12543 & 204585(180303)\\
		& & & dev & 2002 & 25148(21995)\\
		& & & test & 2077 & 25096(21898)\\
		\hline
		\multirow{3}{*}{Estonian (et)} & \multirow{3}{*}{Uralic} & \multirow{3}{*}{EDT} & train & 20827 & 287859(240496)\\
		& & & dev & 2633 & 37219(30937)\\
		& & & test & 2737 & 41273(34837)\\
		\hline
		\multirow{3}{*}{Finnish (fi)} & \multirow{3}{*}{Uralic} & \multirow{3}{*}{TDT} & train & 12217 & 162621(138324)\\
		& & & dev & 1364 & 18290(15631)\\
		& & & test & 1555 & 21041(17908)\\
		\hline
		\multirow{3}{*}{French (fr)} & \multirow{3}{*}{IE.Romance} & \multirow{3}{*}{GSD} & train & 14554 & 356638(316780)\\
		& & & dev & 1478 & 35768(31896)\\
		& & & test & 416 & 10020(8795)\\
		\hline
		\multirow{3}{*}{German (de)} & \multirow{3}{*}{IE.Germanic} & \multirow{3}{*}{GSD} & train & 13814 & 263804(229338)\\
		& & & dev & 799 & 12486(10809)\\
		& & & test & 977 & 16498(14132)\\
		\hline
		\multirow{3}{*}{Hebrew (he)} & \multirow{3}{*}{Afro-Asiatic} & \multirow{3}{*}{HTB} & train & 5241 & 137680(122122)\\
		& & & dev & 484 & 11408(10050)\\
		& & & test & 491 & 12281(10895)\\
		\hline
		\multirow{3}{*}{Hindi (hi)} & \multirow{3}{*}{IE.Indic} & \multirow{3}{*}{HDTB} & train & 13304 & 281057(262389)\\
		& & & dev & 1659 & 35217(32850)\\
		& & & test & 1684 & 35430(33010)\\
		\hline
		\multirow{3}{*}{Indonesian (id)} & \multirow{3}{*}{Austronesian} & \multirow{3}{*}{GSD} & train & 4477 & 97531(82617)\\
		& & & dev & 559 & 12612(10634)\\
		& & & test & 557 & 11780(10026)\\
		\hline
		\multirow{3}{*}{Italian (it)} & \multirow{3}{*}{IE.Romance} & \multirow{3}{*}{ISDT} & train & 13121 & 276019(244632)\\
		& & & dev & 564 & 11908(10490)\\
		& & & test & 482 & 10417(9237)\\
		\hline
		\multirow{3}{*}{Japanese (ja)} & \multirow{3}{*}{Japanese} & \multirow{3}{*}{GSD} & train & 7164 & 161900(144045)\\
		& & & dev & 511 & 11556(10326)\\
		& & & test & 557 & 12615(11258)\\
		\hline
		\multirow{3}{*}{Korean (ko)} & \multirow{3}{*}{Korean} & \multirow{3}{*}{{GSD,Kaist}} & train & 27410 & 353133(312481)\\
		& & & dev & 3016 & 37236(32770)\\
		& & & test & 3276 & 40043(35286)\\
		\hline
		\multirow{3}{*}{Norwegian (no)} & \multirow{3}{*}{IE.Germanic} & \multirow{3}{*}{{Bokmaal,Nynorsk}} & train & 29870 & 489217(432597)\\
		& & & dev & 4300 & 67619(59784)\\
		& & & test & 3450 & 54739(48588)\\
		\hline
		\multirow{3}{*}{Polish (pl)} & \multirow{3}{*}{IE.Slavic} & \multirow{3}{*}{{LFG,SZ}} & train & 19874 & 167251(136504)\\
		& & & dev & 2772 & 23367(19144)\\
		& & & test & 2827 & 23920(19590)\\
		\hline
		\multirow{3}{*}{Portuguese (pt)} & \multirow{3}{*}{IE.Romance} & \multirow{3}{*}{{Bosque,GSD}} & train & 17993 & 462494(400343)\\
		& & & dev & 1770 & 42980(37244)\\
		& & & test & 1681 & 41697(36100)\\
		\hline
		\multirow{3}{*}{Romanian (ro)} & \multirow{3}{*}{IE.Romance} & \multirow{3}{*}{RRT} & train & 8043 & 185113(161429)\\
		& & & dev & 752 & 17074(14851)\\
		& & & test & 729 & 16324(14241)\\
		\hline
		\multirow{3}{*}{Russian (ru)} & \multirow{3}{*}{IE.Slavic} & \multirow{3}{*}{SynTagRus} & train & 48814 & 870474(711647)\\
		& & & dev & 6584 & 118487(95740)\\
		& & & test & 6491 & 117329(95799)\\
		\hline
		\multirow{3}{*}{Serbian (sr)} &
		\multirow{3}{*}{IE.Slavic} & \multirow{3}{*}{SET} & train & 3328 & 74259(74259)\\
		& & & dev & 599 & 11993(11993)\\
		& & & test & 600 & 11421(11421) \\
		\hline
		\multirow{3}{*}{Slovak (sk)} & \multirow{3}{*}{IE.Slavic} & \multirow{3}{*}{SNK} & train & 8483 & 80575(65042)\\
		& & & dev & 1060 & 12440(10641)\\
		& & & test & 1061 & 13028(11208)\\
		\hline
		\multirow{3}{*}{Slovenian (sl)} & \multirow{3}{*}{IE.Slavic} & \multirow{3}{*}{{SSJ, SST}} & train & 8556 & 132003(116730)\\
		& & & dev & 734 & 14063(12271)\\
		& & & test & 1898 & 24092(22017)\\
		\hline
		\multirow{3}{*}{Spanish (es)} & \multirow{3}{*}{IE.Romance} & \multirow{3}{*}{{GSD,AnCora}} & train & 28492 & 827053(730062)\\
		& & & dev & 3054 & 89487(78951)\\
		& & & test & 2147 & 64617(56973)\\
		\hline
		\multirow{3}{*}{Swedish (sv)} & \multirow{3}{*}{IE.Germanic} & \multirow{3}{*}{Talbanken} & train & 4303 & 66645(59268)\\
		& & & dev & 504 & 9797(8825)\\
		& & & test & 1219 & 20377(18272)\\
		\hline
		\multirow{3}{*}{Turkish (tr)} &
		\multirow{3}{*}{Altaic} & \multirow{3}{*}{IMST} & train & 36822 & 37784(36822)\\
		& & & dev & 988 & 10046(9777)\\
		& & & test & 983 & 10029(9797)\\
		\hline
		\multirow{3}{*}{Basque (eu)} &
		\multirow{3}{*}{Language Iasolate} & \multirow{3}{*}{BDT} & train & 5396 & 72974(72974)\\
		& & & dev & 1798 & 24095(24095)\\
		& & & test & 1799 & 24074(24374) \\
		\hline
		\multirow{3}{*}{Persian (fa)} &
		\multirow{3}{*}{IE.Iranic} & \multirow{3}{*}{UPDT} & train & 4798 & 121064(119945)\\
		& & & dev & 599 & 15832(15755)\\
		& & & test & 600 & 16020(15925) \\
		\hline 
	
		\caption{Statistics of the UD Treebanks we used. For language family, ``IE'' is the abbreviation for Indo-European. ``(w/o) punct'' means the numbers of the tokens excluding ``PUNCT'' and ``SYM''.}
		
		\label{tab:udstat}
	\end{longtable}
\end{small}

\clearpage
\setlength{\tabcolsep}{2pt}
\begin{table}[h]
    \scriptsize
    \centering
    \begin{minipage}{0.5\textwidth}
        \begin{tabular}{l| c} 
             \toprule
            {\bf Language Family} & {\bf Languages} \\ 
             \hline
        		Afro-Asiatic & Arabic (ar), Hebrew (he)\\
        		\hline
        		Austronesian & Indonesian (id)\\
        		\hline
        		\multirow{2}{*}{IE.Germanic} & Norwegian (no), Danish (da), Dutch (nl),  \\
        		& English (en), German (de),  Swedish (sv)\\
        		\hline
        		IE.Indic & Hindi (hi)\\
        		\hline
        		IE.Altaic & Turkish (tr)\\
        		\hline
        		\multirow{2}{*} {IE.Romance} & Catalan (ca), French (fr), Portuguese (pt),\\
                &  Italian (it), Romanian (ro), Spanish (es)\\
        		\hline
        		\multirow{2}{*} { IE.Slavic} & Bulgarian (bg), Croatian (hr), Czech (cs), Polish (pl), \\
        		& Russian (ru), Slovak (sk), Slovenian (sl), Serbian (sr)\\
        		\hline
        		Japanese & Japanese (ja)\\
        		\hline
        		Korean & Korean (ko)\\
        		\hline
        		Sino-Tibetan & Chinese (zh)\\
        		\hline
        		Uralic &  Finnish (fi)\\
        		\hline
        		Iranic & Persian (fa) \\
        		\hline
        		Isolate & Basque (eu)\\
        % 		\hline
             \bottomrule
        \end{tabular}
        \caption{\textbf{\scriptsize{ The selected languages from UDTreebank 2.2,used in our cross lingual POS tagging, grouped by language families. ``IE'' is the abbreviation of Indo-European.
         }}}
        \label{table:case-rcp}
    \end{minipage}
\end{table}

\clearpage

\subsection{POS Taggong Dataset for Domain Transfer}
\label{app:B}
\setlength{\tabcolsep}{2pt}
\begin{table}[h]
% \scriptsize
\footnotesize
\centering
    \begin{minipage}{\textwidth}
        \begin{tabular}{l l l c c c c} 
            \toprule
     
            {\bf Domain}  & {\bf WSJ } & {\bf Emails } & {\bf Newsgroups}  & {\bf Answers} & {\bf Reviews}  & {\bf Weblogs} \\ 
            Train/Dev/Test & 2976/1336/1640 & 4900/2450/2450 & 2391/1196/1195 & 3489/1745/1744 & 3813/1907/1906 & 2031/1016/1015 \\
            \bottomrule
        \end{tabular}
        
        \caption{\textbf{{ Data Statistics of SANCL 2012 shared task dataset \cite{sancl2012}} }}
        
        \label{table:case-rcp}
    \end{minipage}
\end{table}

\clearpage
% \newpage
\subsection{Modified GLUE NLI Task for Domain Transfer}
\label{app:C}
For NLI, we consider the 2-class classification (e.g., entailed or not) corpora used in \citet{bert-domain} that is made upon  modification on 4 Glue benchmark \cite{glue} problems: SNLI, MNLI, QNLI, and QQP. 
We split the MNLI training set into a corpora of ``fiction'', ``slate'', ``govt.'', ``travel'', and ``telephone''  as in \citet{MNLI} and always include them in the source corpora for all target domains. Here, As the annotations for GLUE test sets are publicly unavailable, for each target domain, we consider the original dev set as pseudo test set and randomly select 2k  instances from training set for parameter tuning (i.e., pseudo dev set). For MNLI as target, we have two original dev set. Hence, we take the 2k instances from matched dev set as pseudo dev set and consider the miss-matched corpus as pseudo test set.
Therefore, in zero-shot setting, the number of source corpora for target MNLI is 8, and for others it is 7. 
\setlength{\tabcolsep}{2pt}
\begin{table}[h]
% \scriptsize

% \centering
% \hspace{-10pt}
    \begin{minipage}{0.5\textwidth}
        \begin{tabular}{l l l c} 
            \toprule
     
            {\bf Task Category } & {\bf Dataset } & {\bf Train Size}  & {\bf Dev Size } \\ 
            \midrule
            Natural Language Inference  & SNLI & 510,711 & - \\
            \midrule
             & MNLI-Fiction & 77348 & - \\
             & MNLI-Travel & 77350 & -\\
            Multi-Genre & MNLI-Slate & 77306 & - \\
             Natural Language Inference& MNLI-Government & 77350 & - \\
             & MNLI-Telephone & 83348 & - \\
            & MNLI-Mismatched & \quad  - & 9,832\\
            \midrule
            Answer Sentence Selection & QNLI & 108,436 & 5,732\\
            \midrule
            Paraphrase Detection & QQP & 363,847 & -\\
             \bottomrule
        \end{tabular}
        
        \caption{\textbf{{ Data Statistics of Glue NLI tasks. We report the performance on the full dev set and to tune all the models, randomly select 20\% examples from it. } }}
        
        \label{table:case-rcp}
    \end{minipage}
\end{table}

\clearpage

\subsection{Sentiment Analysis Dataset for Domain Transfer}
\label{app:D}
For sentiment analysis, following \citet{ijcaibooks}, we use the multi-domain sentiment datasets released by \citet{liu-etal-2017-adversarial} which has several additional domains than a popular sentiment analysis dataset \citet{blitzer-etal-2007-biographies}.
For each domain, we use the same test set as in \citet{liu-etal-2017-adversarial}. However, as train and dev data are released together, we simply consider the first section of this combined set as the train set and the last section as dev set. Statistics of the 14 domains in this dataset considered in our experiments are reported in Appendix.
\setlength{\tabcolsep}{2pt}
\begin{table}[h]
% \scriptsize

    \begin{minipage}{0.5\textwidth}
    \centering
        \begin{tabular}{l l c c c c} 
            \toprule
            Dataset & Train & Dev & Test & Avg length\\
            \midrule
                Books & 1400& 200 & 400 & 159 \\
                Electronics & 1398 & 200 & 400 & 101 \\
                DVD & 1400 & 200 & 400 & 173 \\
                Kitchen & 1400 & 200 &400 &89\\
                Apparel & 1400 &200 &400 &57 \\
                Camera &1397 &200& 400& 130\\
                Health &1400& 200 &400& 81\\
                % Music 1400 200 400 136
                Toys &1400 & 200 &400& 90\\
                Video &1400 &200& 400 &156\\
                Baby & 1300 &200 &400 &104 \\
                Magazine &1370& 200& 400& 117\\
                Software &1315& 200 &400 &129\\
                Sports & 1400 & 200 &400 &94\\
                % IMDB 1400 200 400 269
                MR &1400& 200& 400 &21\\

            \bottomrule
        \end{tabular}
        
        \caption{\textbf{{ Data Statistics of multi-domain sentiment dataset} }}
        
        \label{table:case-rcp}
    \end{minipage}
\end{table}

\subsection{XNLI Results}
\label{app:E}
\setlength{\tabcolsep}{3pt}
\begin{table}[h]
\scriptsize
\centering
% \begin{minipage}{0.5\textwidth}
 \begin{tabular}{l| l| c| c| c| c| c| l |  l| c| c| c| c| c| l | c} 
 \toprule
{\bf Model }   &  {\bf es } & {\bf de } & {\bf el } & {\bf bg } & {\bf ru } & {\bf tr } & {\bf ar } & {\bf vi } & {\bf th } & {\bf zh } & {\bf hi } & {\bf sw } & {\bf ur } & {\bf en } & {\bf fr}\\
\midrule
% \hline 

All Sources & 77.88 & 71.82 & 72.23 & 74.03 &  73.59 & 65.21 & 68.94  & 74.39 & 60.10  & 74.69 & 67.31 & 52.67  & 64.37 & 82.65 &  77.03
\\
Baseline-r& 77.88 & 71.82 & 72.23 &  74.69 & {\bf 74.53} & {\bf 65.85} & 68.68 & 75.03 & 60.10 & 74.69 & 66.69 & {\bf 52.97} & 63.69 & 82.65 &77.03
\\
Baseline-s & 77.88 & 71.82 & 72.23 &   73.23 &   73.73  & {65.67} & {68.36} & {74.11} & 60.10 &   74.69 & {67.07} & { 52.59} & {63.31} & 82.65 & 77.03
\\
Ours & 77.88 & 71.82 & 72.23 &  {\bf 74.95} &   73.85  & { 65.63} & {\bf 69.24} & {\bf 75.71} & 60.10 &   74.69 & {\bf 67.78} & { 52.73} & {\bf 64.67} & 82.65 & 77.03
\\
 \bottomrule
 \end{tabular}
 \caption{\textbf{{ Cross-lingual results on the XNLI  test sets.
 }}}
 \label{table:case-rcp}
%  \end{minipage}
\end{table}

All model performances are same when selecting all source corpora as potential.

\cleardoublepage
\subsection{How good is the approximation?}
\label{app:F}
We consider three different datasets: XNLI (target: 'es'), Modified GLUE NLI dataset(target: 'MNLI-mm'), and SANCL 2012 shared task for POS tagging (target: 'WSJ'). We use the corresponding full size source tasks except for the extremely large XNLI in which we randomly sample half of each source task (180k instances) and compute the Shapley value adopting  the source code released by \citet{james-data-shapley}. Then, on XNLI dataset we consider sample size 50k, on GLUE 20k, on SANCL 2012 sahred task 2k for each source task. Then we use Algorithm 1 (in the main paper) with tuned initial score to compute the approximate data Shapley value. Instead of full convergence, we do early stop by setting the Shapley value nepochs to 10, 50, 30 on XNLI, GLUE, and SANCL datsets respectively. 

\clearpage
\subsection{Shapley Value Computation Time with/out different Factors: }
\label{app:G}

We consider two example problem to transfer both language and domain: (i) UDPOS tagging for language transfer (ii) modified GLUE NLI for domain transfer. We consider the ``initial score'' to {\it All Sources/2} and $\mathcal{R}$ ; $nepoch$ to 30, and 50 for these two respective target task, for the data Shapley computation as in Algorithm 1, we then  switch different factors as in reported in Table 6 (in the main paper)  as record the corresponding Shapley value computation time. 

\clearpage
\subsection{Approximate Shapley Value with Different Seeds: }
\label{app:H}
\begin{figure}[h]
\hspace{-0.2cm}
\centering
\includegraphics[width=15cm, height=5cm ]{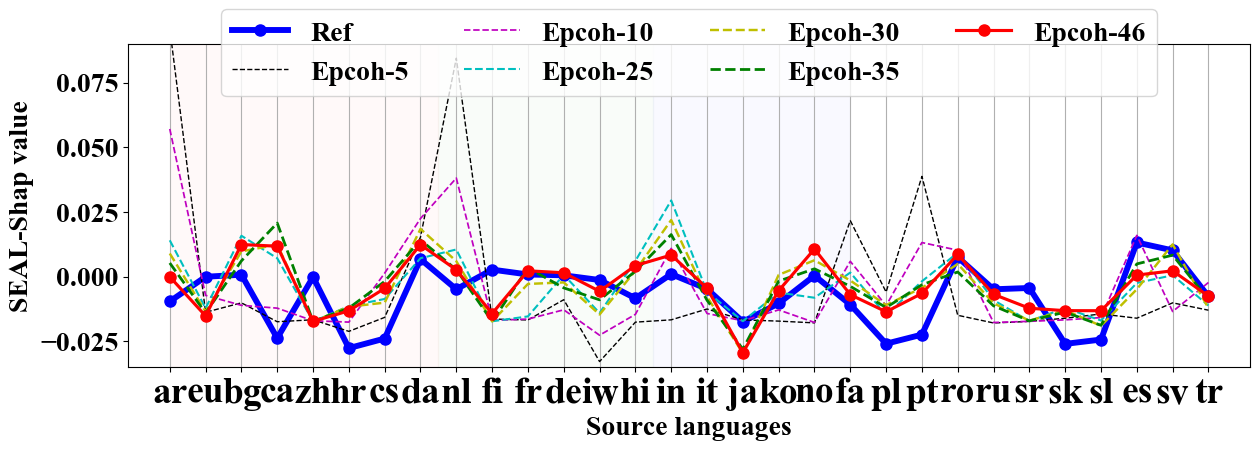}
\caption{ \mym value with two different seeds.} 
\vspace{-0.2cm}
  \label{fig:shap-diff-seed}
\end{figure}

In Figure \ref{fig:shap-diff-seed}, we plot the \mym  value  w.r.t same threshold ($\theta_{k} = 0$) by different seeds.

\clearpage
\subsection{Adding sources according different source ranking/selection methods}
\label{app:I}
\begin{figure}[h]
\hspace{-0.2cm}
\centering
\includegraphics[width=7.5cm, height=5cm ]{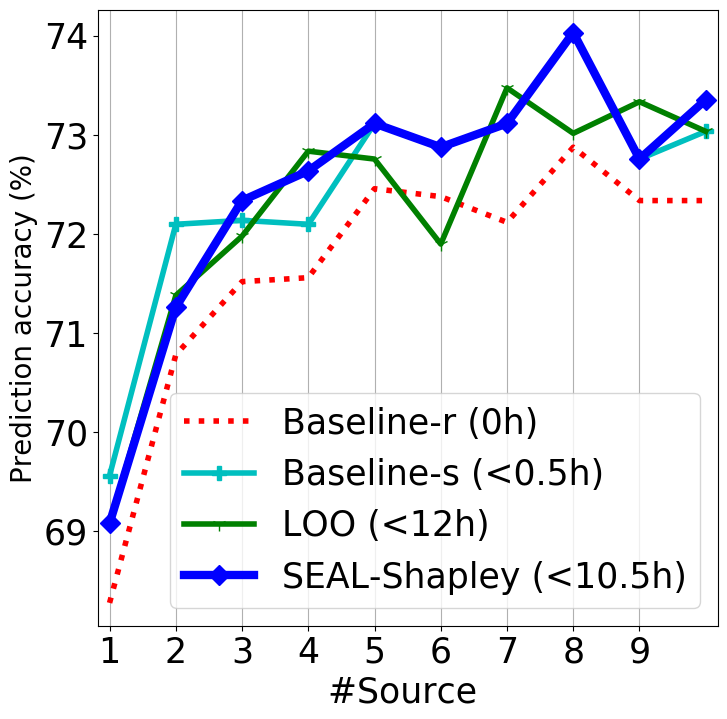}
\caption{ Performance using top-k sources as per dfifferent source ranking/selection methods. (Task: XNLI, target: vi. Red colored line denotes Random or Baseline-S. } 
\vspace{-0.2cm}
  \label{fig:shap-diff-selection-method-electronics}
\end{figure}
\begin{figure}[h]
\hspace{-0.2cm}
\centering
\includegraphics[width=7.5cm, height=5cm ]{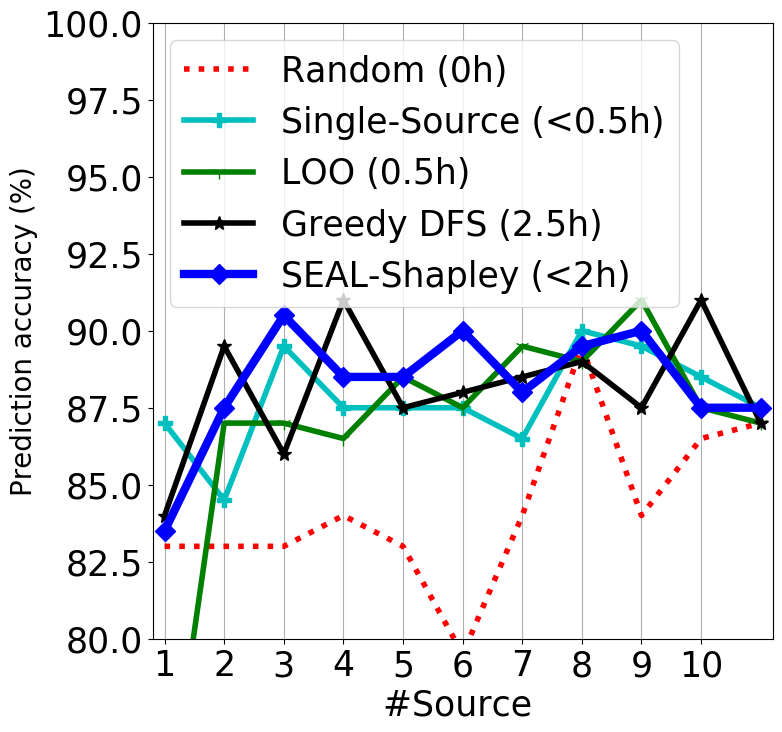}
\caption{ Performance using top-k sources as per different source ranking/selection methods (Task: Cross-domain Sentiment Analysis, Target: Electronics.) Red colored line denotes Random or Baseline-S.} 
\vspace{-0.2cm}
  \label{fig:shap-diff-selection-method-electronics}
\end{figure}
\begin{figure}[h]
\hspace{-0.2cm}
\centering
\includegraphics[width=7.5cm, height=5cm ]{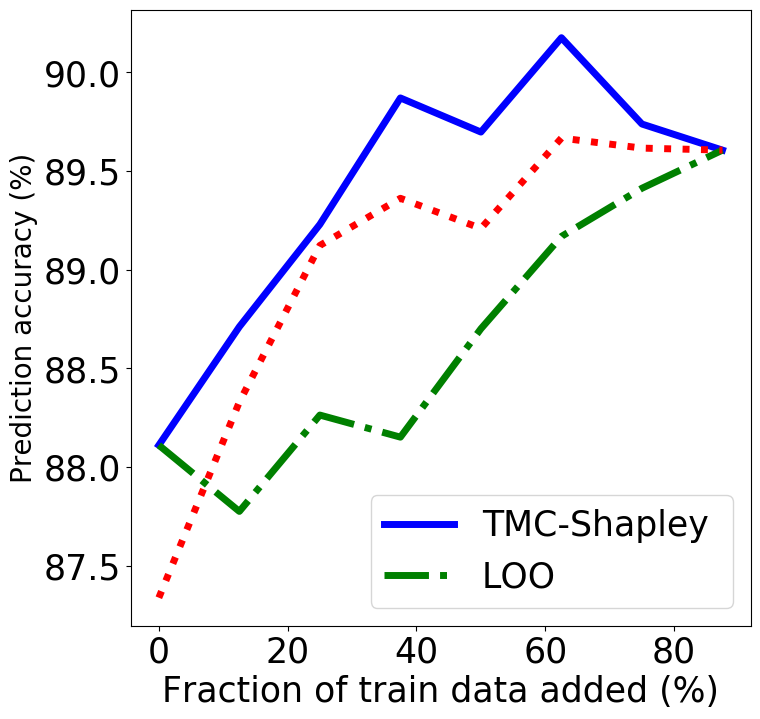}
\caption{ Performance using top-k sources as per dfifferent source ranking/selection methods. (Task: Cross-domain NLI, target: MNLI-mm). Red colored line denotes Random or Baseline-S. } 
\vspace{-0.2cm}
  \label{fig:shap-diff-selection-method-electronics}
\end{figure}

\newpage

We consider a Bert model with certain model parameters. Then using the corresponding training and development dataset we  compute \mym values by adjusting the $\rho$, compute the ranks according to Baseline-s, Baseline-r, language distance from the target language etc., We consider the top-3 sources to compare. We also consider  the top-3 sources in a greedy depth first search approach. Ours get consistent increase and best performance using top-3 sources. Here we plot figure with more top-3 sources. For large datasets, we do not plot the greedy DFS here as it takes extremely long time to compute due to the fact that the DFS search branches rarely overlap for different targets.

% \begin{figure}[h]
% \hspace{-0.2cm}
% \centering
% \includegraphics[width=5cm, height=3cm ]{image/Adding_sources_on_VI.png}
% \caption{ \mym value with two different seeds.} 
% \vspace{-0.2cm}
%   \label{fig:shap-diff-selection-method-electronics}
% \end{figure}

\clearpage

\subsection{Classifier and Data Preprocessing: }
\label{app:J}
As for the underlying machine learning classifier, in our experiments on the domain transfer problem, we consider the BERT based cased model \cite{devlin2019bert} except for POS tagging. For POS tagging, we consider the the state-of-the-art BiLSTM based Flair framework \cite{flair-18}. As for language transfer problem,  we consider a generic state-of-the-art classifier: the multi-linigual version of BERT based cased model. For all bert models, we adopt Transformers implementation \cite{HuggingFacesTS}. Number of model parameters BERT model 10 million parameters.
For each task, no preprocessing is performed other than the tokenization of words into subwords with WordPiece except for cross-lingual POS for which we use an oppen-sourced multilingual preprocessing toolkit\footnote{\url{github.com/huggingface/transformers/tree/master/examples/token-classification}} to remove ``strange control character" tokens. 
Following \citet{mbert}, we also limit subwords sequence length to 128 to fit in a single GPU for all tasks. For all tasks, we use the accuracy metric. 
% \paragraph{Baselines: } 
\clearpage
\subsection{Hyper-parameters Tuning: } 
\label{app:K}
For the small multi-domain sentiments analysis dataset, we do a full search of the combination of learning rate, batch size and number epochs up to 5. For all other large scale datasets, we perform a greedy search. We first find the best combination of learning rate and batch size. Then we tune the number of epochs. For the extremely large XNLI, in which for any target task the multi-source training data size is $\sim5.5$M, we tune only when our framework select a smaller subset of the source corpora  for learning rate in $\{3\times 10^{-5}, 5\times 10^{-5}\}$, batch size $\{32\}$ and epochs within $50$k steps (i.e., no more than 3 epochs). On XNLI, when our framework selects all source corpora as potential, we do not further tune the hyper-parameters both baselines have the same training set as \mym. Hence,  we report the result using a default learning rate $5\time 10^{-5}$, batch size 32.   All test results reported in this paper are performed on the corresponding test set\footnote{for GLUE NLI, pseudo test set } using a single gpu. All Shapley value calculations were performed on multi-gpus. After transfer source selection, all models for XLNI, and UDPOS, are trained, and tuned on multi-gpu distributed system and for for SANCL 2012 POS tagging, and mulit-domain sentiment analysis datasets single gpu is used. As for modified GLUE NLI dataset, both single gpu and distributed system is used. For UDPOS significance test, we use the default num\_samples 10k, except for Polish we use 3k. As for Flair, the system does not support saving the prediction options, getting the model prediction even from a trained BiLSTM model  is time consuming. Hence, we sample for no more than 50 times for SANCL 2012 dataset significance test.   
% All the Shapley value computation runs are computed over runs from a single seed except when the number of sources are high like on UDPOS, and XNLI. Even for them at most 3 seeds are used.
For Flair framework, after preliminary verification, we follow their configuration suggested for best performance on English Penn treebank POS tags\footnote{\url{github.com/flairNLP/flair/blob/master/resources/docs/EXPERIMENTS.md}} and tune each model up to 150 epochs with patience 4. All the approximated Shapley values (\mym values) are computed within $nepoch$ 30 and only for UDPOS and XNLI dataset, multiple seeds ($<3$)  are used. For UDPOS  $nepoch$ within 30 or 46, for XNLI $nepoch$ within 10 or 20.  For any target task $V_k$, the corresponding threshold Shapley value $\theta_k$ is chosen in $\{1\times 10^{-2}, 1\times 10^{-3}, 5\times 10^{-3}\}$, initial scores $\rho$ in $\{\mathcal{R}, \mathcal{N}, 0.5, {\textit{ All sources}/2, \textit{ All Sources}}, \mu  \}$  where $\mathcal{R}$ is a  random baseline model performance (i.e., randomly initialized model performance); given the total number of sources $n$, $\mathcal{N} = \frac{n-1}{n} \times Score(C_{D_j}, V_k)$; $\mu = mean(\{\textit{ All Sources}\} \cup \{Score(C_{D_j}, V_k) \forall D_j \in D\})$,  and $D$ is all source tasks. This means we also tune \mym value as the mean of a combination of \mym values, leave one out values, and single source transfer values. For Shapley value computation, for multiple seeds run like on cross-lingual POS tagging and  XNLI, seed 42, 43 is used.  All the hyper-parameter tuning is done with default seed in the open-sourced Transformer implementation\footnote{\url{github.com/huggingface/transformers/}} which is 42.  All the \mym values are calculated using single seed.  Only for plotting Figure 6 in main paper, on cross-lingual POS tagging for target English  two different seeds are used. The blue curve in Figure 6, and the results in Table 2 are using the same seed and all other plots uses the other seed.   All the parameter configuration and the dev set performance will be reported here upon acceptance. All computations are performed on gpus; in general using  (4,8,1) \#gpus. Note that while tuning, if there is no $\theta$ for which the corresponding subset of sources (i.e., $\subset \mathcal{D}$) achieves better result than using all of $\mathcal{D}$, then we select the set of all  sources  $\mathcal{D}$ assuming each source is contributing positively.

\clearpage

\subsection{Training a Direct Source Selection Ranker using \mym}
\label{app:L}
\begin{figure}[h]
\hspace{1.2cm}
\centering
\includegraphics[width=15cm, height=7cm ]{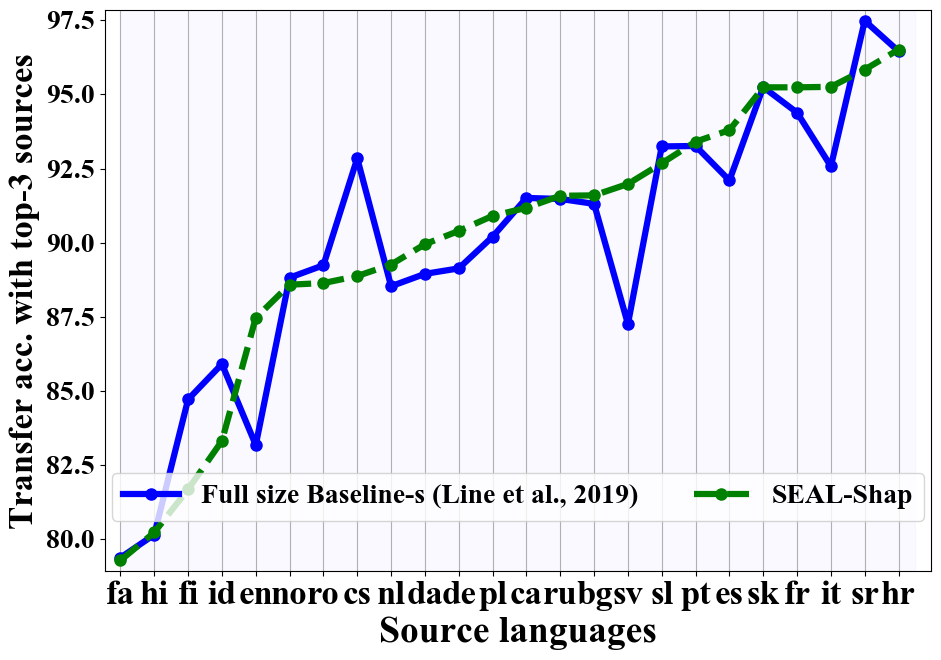}
\caption{ Transfer performance with the top-3 sources} 
\vspace{-0.2cm}
  \label{fig:linetal_ori_vs_seal_shap}
\end{figure}
In addition to the sampled single source performances {\it Baseline-s} in the main paper, here we also compare smapled \mym with the full size (i.e., no sampling) single source performance.
\citet{neubig-choosing} with full size {\it Baseline-s} results are found using the original ranker realised in \citep{neubig-choosing}.   In most cases ours outperforms \cite{neubig-choosing}  such as 'hr', 'de', 'da', 'nl', 'en', 'fr', 'he', 'it', 'es', 'sv'. However, the margin is small and also there are multiple cases where \citet{neubig-choosing} outperforms ours such as 
'ar', 'cs', 'zh',
'id', 'fi', 
'ja', 'ko',
'sr'.

\clearpage

\subsection{Dev set Results}
\label{app:M}
In Table \ref{tab:dev-lang-udpos}, we report the dev set result for cross-lingual POS tagging. For French, German, Hebrew, Slovenian, we need 

\begin{table}[h]
	\centering
\resizebox{0.5\linewidth}{!}{ 
\scriptsize
	\begin{tabular}{@{}c@{ }|c@{ }||c@{ }|c@{ }|c@{}}
		\hline
		{\bf Lang} & {\bf All Sources} & {\bf  Baseline-s} & {\bf \mym}  \\
		\hline
		%=====
            en & 85.21 & 87.39 &
            {\bf 88.50}
            \\
            % \hline
            % \multicolumn{7}{l}{Target Languages} \\ 
            % \hline
            % Norweian
            no  & 90.05 & 90.05 &
            90.05
            \\
            % Swedish
            % sv &    92.97+92.36 & 93.05 & 
            % {\bf 92.94+93.37}
             sv &    {\bf 93.39} & 93.27 & 
            93.18
            \\ 
            % French
            fr & 95.52 & 95.68 &
            {\bf 95.71}
            
          \\
        %   Portuguese
            pt & 94.55 & 94.73 & { \bf 94.77}
            \\
            % danish
            da & 90.19 & 90.27 & {\bf 90.42}
            \\
            % Spanish
            es & 94.11 & 94.04 &
            { \bf 94.16}
            \\
            % italian
            it & 96.83 & 96.56&
            {\bf 96.89}
            \\
            % croatian
            hr &  96.36 & 96.36  & 96.36 
        
            \\
            % catalan
            ca &  92.58 & 92.39 & {\bf 92.92}
            
            \\
            % polish
            pl & {\bf 91.64} & 91.40 & 91.62
            % for this result 1500, 3000 num_smaples used in t-test
            \\
            % Slovenian
            sl &  93.35 & 93.56 &
            {\bf 93.45}
            \\
            % dutch
            nl &  91.47 & 91.55 & 
            {\bf 91.55}
            \\
            % Bulgarian
            bg & 92.26 & 92.26 & 
            92.26 
            \\
            % russian
            ru & 92.87 & 92.79
            & {\bf 92.92}
            \\
            % German
            de &  91.42 & 91.65 & 
            91.42
            % both better than ALL sources
            \\
            % Hebrew
            he & 77.09 & 76.16 &
            {\bf 77.30}
            \\
            % czech
            cs &  94.56 & 93.14 & 
            {\bf 94.74}
            \\
            % romanian
            ro &  90.41 & 90.41 & 90.41
            \\
            % slovak
            sk &  96.38 & 96.33 &
            {\bf 96.42} & 
            \\
            % Serbian
            sr &  97.18 & 97.27 &
            {\bf 97.35}
            \\
            % indonesian
            id & 83.98 & 84.63 & 
            {\bf 85.58}
            \\
            % finnish
            fi &  {\color{blue}{87.24}} & 87.26 & { 87.26}
            \\
            % chinese
            zh &   71.31 & 71.31 &
            71.31
            \\
            % arabic
            ar & 79.18 & {79.18} & 79.18 
            \\
            % Korean
            ko & 63.58 &  63.76 &
             {\bf 64.31}
            \\
            % hindi
            hi &  80.69 & 80.16 &
             {\bf 82.78}
            \\
            % Japanese
            ja &  69.28 & 69.72 & 
            {\bf 69.93}
            \\
            % turkish
             tr &   78.43 & 78.43 & 
             78.43
            \\
            eu &  80.90 & 80.90 &
            80.90
            \\
            fa &  82.37 & 81.67 &
            {\bf 82.74}
            \\
            \hline
            % {\bf Average} & - & 87.17 & 87.21 & {\bf 87.62 }
            \\
		%===========
		\hline
	\end{tabular}
}
	\caption{
	\label{tab:dev-lang-udpos}
{ Dev set results on cross-lingual POS tagging.
}
		}
\vspace{-18pt}
\end{table}

\cleardoublepage

\clearpage

\subsection{Full Cross-lingual POS Tagging Results }
\label{app:N}
\begin{table}[h]
	\centering

\resizebox{0.5\linewidth}{!}{ 

	\begin{tabular}{|@{}c@{ }|c@{ }||c@{ }|c@{ }|c@{}|c@{}|c@{}|c@{}|}
		\hline
		{\bf Lang} & {\bf en} & {\bf All Source} & {\bf  Baseline-r} & {\bf SEAL-Shap}  & {\bf Baseline-s}\\
% 		 &  &  &  & {\bf Shap}  &  \\
		\hline
		%=====
            en & - & {82.71} & 86.32 &
            {\bf 88.55$^{*\$}$} & 86.39
            \\
            % \hline
            % \multicolumn{7}{l}{Target Languages} \\ 
            % \hline
            % Norweian
            no & - & 90.06 & 90.06 &
            90.06  & 90.06
            \\
            % Swedish
            sv &  83.6 & 93.26 & 
            % 93.31 & 
            93.26 & 93.26
            &
            % {\bf 93.37} 
            93.26
            \\
            % French
            fr & - & 94.60 & 94.63 &
             94.79 & {\bf 94.83}
            
          \\
        %   Portuguese
            pt & 82.1 & 94.33 & 94.33 &
          94.33 & 94.33
            \\
            % danish
            da & 88.3 & 88.94 & 89.30 &
             {\bf 89.47$^{*}$} & 89.23
            \\
            % Spanish
            es & 85.2 & 93.15 & 93.00 &
            { \bf 93.21$^{\$}$} &93.04
            \\
            % italian
            it & 84.7 & 96.58 & 96.43&
            % {\bf 
            96.67
            % $^{\$}$} 
            & {\bf 96.71}
            \\
            % croatian
            hr & - & 96.60 & 96.60&
          96.60 & 96.60
            \\
            % catalan
            ca &  - & 91.54 & 91.64 &
            {\bf 92.08$^{*\$}$} & 90.78
            \\
            % polish
            pl & 86.9 & 91.61 & 
            % 91.48 &
            91.61 &
            91.61 &
            91.61
            % {\bf 91.63$^\$$}
            
            % for this result 1500, 3000 num_smaples used in t-test
            \\
            % Slovenian
            sl & 84.2 & 93.28 & 93.50 &
            {\bf 93.52$^*$} & 92.89
            \\
            % dutch
            nl & 75.9 & 90.10 & 90.19 & 
            {\bf 90.26} & 90.14
            \\
            % Bulgarian
            bg & 87.4 & 92.93 & 92.93 & 
             92.93 & 92.93
            \\
            % russian
            ru & - & 92.98 & 92.91
            & {\bf 93.13$^{*\$}$} & 92.71
            \\
            % German
            de & 89.8 & 90.79 & { 91.07} & 
            91.06 & {\bf \bf 91.44}
            % both better than ALL sources
            \\
            % Hebrew
            he & - & 76.67 & 75.75 &
            {\bf 76.73$^\$$} & 75.43
            \\
            % czech
            cs & - & 93.89 & 93.04 & 
            {\bf 94.81$^{*\$}$} & 93.94
            \\
            % romanian
            ro &  84.7 & 89.97 & 89.97 &
             89.97 & 89.97
            \\
            % slovak
            sk &  83.6 & 95.68 & 95.62 &
            {\bf 95.81} & 95.53
            \\
            % Serbian
            sr &  - & 97.55 & 97.47 &
            {\bf 97.58} & 97.43
            \\
            % indonesian
            id &  - & 84.10 & 85.23 & 
            {\bf 85.97$^{*\$}$} & 85.50
            \\
            % finnish
            fi &   - & {\bf  87.13} & 86.89 &
            87.05 & 86.86
            \\
            % chinese
            zh &  - &  71.31 & 71.31 &
            71.31 & 71.31
            \\
            ar & - & {80.07} & 80.07 & 
            80.07 & 80.07 
            \\
            % Korean
            ko &  - & 63.59 & {\bf 64.27} &
             64.19 & 63.77
            \\
            % hindi
            hi &  - & 81.49 & 80.27 &
             {\bf 82.41$^{*\$}$} & 79.94
            \\
            % Japanese
            ja &  - &  66.86 & 65.99 & 
            {\bf 67.81$^{*\$}$} & {67.71}
            \\
            % turkish
             tr &  - &  78.43 & 78.43 & 
             78.43 & 78.43 
            \\
            eu &  - &  81.18 & 81.18 &
            81.18 & 81.18 
            \\
            fa & 72.8 & 81.03 & 80.69 &
            % {\bf
            81.79
            % $^{*\$}$} 
            & {\bf 82.37}
            \\
            \hline
            {\bf Average} & - & 87.17 & 87.21 & {\bf 87.62 }
            \\
		%===========
		\hline
	\end{tabular}
}
	\caption{
	\label{tab:lang-udpos_full}
{ Performance on universal POS tagging (test set) when using each of language as the target language and the rest as source languages .  '*'  and `\$' denote \mym model is statistically significantly outperforms
{\it All Sources} and {\it Baseline-s } respectively using paired bootstrap test with p $\leq$ 0.05. `en' refers to the best single source ('en') results, reported in \citet{mbert}. %Same accuracy for all models indicates all source tasks are selected (i.e., $Z_k = D$). 
}
		}
% \vspace{-28pt}
\end{table}

 All model performances are same when selecting all source corpora as potential. (See line 2 in Table \ref{tab:dev-lang-udpos}  and Table \ref{tab:lang-udpos_full}).
 
 \clearpage

 \subsection{Full Cross-domain Sentiment Analysis Results }
\label{app:O}
 \setlength{\tabcolsep}{3pt}
\begin{table*}[h]
\label{tab:UPOS}
\centering
\resizebox{\linewidth}{!}{ 
\begin{tabular}{l| l| c| c| c| c| c| l |  l| c| c| c| c| c| l | c  } 
 \toprule
{\bf Model }   &  {\bf books} & {\bf kitchen} & {\bf dvd} & {\bf electronics} & {\bf apparel} & {\bf camera} & {\bf baby} & {\bf health} & {\bf magazines} & {\bf   MR} & {\bf software} & {\bf video} & {\bf toys} & {\bf sports} &{\bf Avg}
\\
 \midrule
\citet{ijcaibooks} &  87.3 & 88.3 &  88.8 & 89.5 & 88.0 & 90.3  & 90.3 & 91.0 & 88.5 & 76.3 &  90.8  &  91.3 & 90.3& 90.5 & 82.16
\\
\midrule
% \hline 
All Sources & {\bf 87.3} & 90.3 & 88.3 & 90.8 & { 91.0} &  91.5 & 92.3 & 92.0 & 90.5 & 79.3 & 90.3 & 85.3 & 91.3 & 90.5 & 89.33
\\
Baseline-r & 87.0 & 90.5 & 87.3 & 90.8  & {91.0} 
% 90.3
&   91.5 & 91.8 & 92.0 & 90.5 &  78.8 & 90.0 & 84.8 & 91.3 & 90.5 & 89.08
\\
Baseline-s & 86.8 & 89.8 & 87.0 & 90.8  & {91.0} 
&   91.5 & {\bf 92.5}  & 92.0 & 90.5 &  77.5 & 90.0 & 84.8 & 91.3 & 90.5 & - \\
\mym & {\bf 87.3} & {\bf 90.8} & {\bf 88.8} & 90.8 & { 91.0}   & 91.5 & {\bf 92.5} & {92.0} & 90.5 & {\bf 79.5}& 90.3 & 87.8 & 91.3 & 90.5 & {\bf 89.76}
\\
 \bottomrule
 \end{tabular}
 }
 \caption{ Cross-domain Transfer performance on multi-domain sentiment analysis dataset  \cite{liu-etal-2017-adversarial}. \citet{ijcaibooks} leverages unlabelled data from the target domain. 
 }
 \label{tab:domain-mtl-senti}
\end{table*}

\clearpage
\subsection{\mym values for two similar targets }
\label{app:P}
 \begin{figure}[h]
\vspace{-0.2cm}
\centering
\includegraphics[width=7.8cm, height=3.8cm]{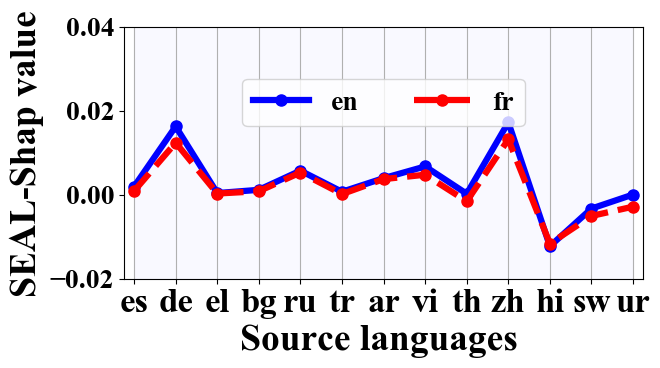}
\vspace{-0.3cm}
\caption{
% \vspace{1.5cm}
{ Similar \mym value curvature  of two close language English (``en'') and French (``fr'') on cross-lingual NLI.  }
\vspace{-0.3cm}
}
  \label{fig:two-sim-xnli}
\end{figure}

\clearpage
\subsection{Interpreting Source Shapley Values in Cross-domain NLI}
\label{app:Q}
 \begin{figure}[h]
% \vspace{-0.18cm}
\centering
\includegraphics[width=7.8cm, height=3.9cm]{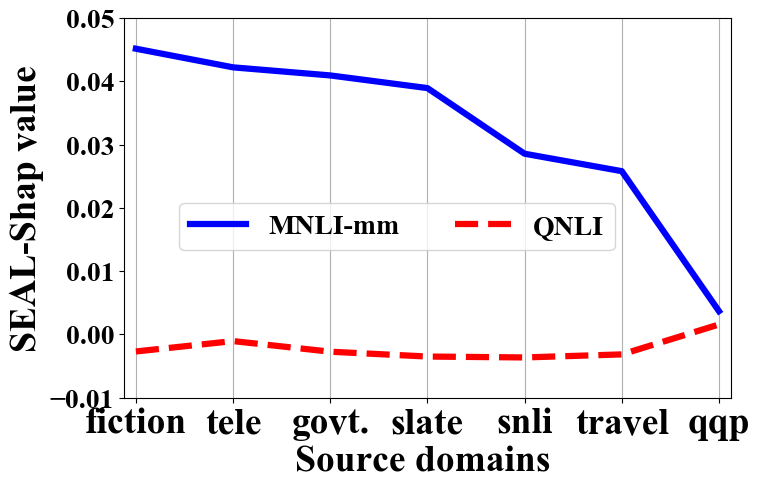}
% \vspace{-0.3cm}
\caption{
% \vspace{1.5cm}
{ \mym value on cross-domain NLI, referring to  relative contribution of source domains. For target domain MNLI-mm, source  domain QQP has the lowest contribution, whereas for target domain QNLI, source domain QQP has the highest contribution. }
% \vspace{-1cm}
}
  \label{fig:two-sim-nli}
\end{figure}

\clearpage

\subsection{Data Statistics}
\label{app:datasatat}
\setlength{\tabcolsep}{5pt}
\begin{table}[h]
% \begin{minipage}{0.5\textwidth}
% \scriptsize
\centering
\resizebox{\linewidth}{!}{ 
 \begin{tabular}{l| l| c| c| c } 
 \toprule
% \hline
   {\bf Transfer} & {\bf Task }  &{\bf Dataset } & {\bf  \#target} & {\bf \#source}  \\
%   &\multirow{2} {*}{$ \lvert \mathcal{D} \rvert $ or $(m)$ }
%   \\
    %  &   &  &  {\bf $(n)$ & {\bf for  each target} & }
%   \\
% \midrule
\hline
\multirow{2}{*}{Language} & POS tag & UD Treebank & 31 & 30 \\
& NLI & XNLI & 15 & 14 \\
\hline
\multirow{3}{*}{Domain} & POS tag & SANCL 2012 & 6 & 5 \\
& NLI & mGLUE & 4 & 7+ \\
& Sentiment Ana. & mlt-dom-senti  & 14 & 13  \\
 \bottomrule
% \hline
 \end{tabular}
 }
 \caption{{ Task statistics. \#sources are for each target. In (m)odified GLUE,  \#sources is 8 for target MNLI, and 7 otherwise. ``mlt-dom-senti'' refers to \citet{liu-etal-2017-adversarial}. 
 }}
 \label{tab:stat-task}
%  \end{minipage}
 \vspace{-1.5em}
\end{table}

\clearpage
\subsection{Number of Sources Selected}
% \label{sec:num-souurce-selection}
\label{app:S}
\begin{table}[h]
	\centering
	\begin{tabular}{l | c}
		\hline
		{ Lang} & {\#Sources Selected} \\
		\hline
		%=====
            en & 9
            \\
            % \hline
            % \multicolumn{7}{l}{Target Languages} \\ 
            % \hline
            % Norweian
            % no  & 30
            % \\
            % Swedish
            %  sv &   30
            % \\ 
            % French
            fr & 29
          \\
        %   Portuguese
            % pt & 
            % \\
            % danish
            da & 29
            \\
            % Spanish
            es & 27
            \\
            % italian
            it & 26
            \\
            % croatian
            % hr &  
        
            % \\
            % catalan
            ca &  25
            
            \\
            % polish
            % pl & 
            % for this result 1500, 3000 num_smaples used in t-test
            % \\
            % Slovenian
            sl &  29
            \\
            % dutch
            nl &  28
            \\
            % Bulgarian
            % bg & 28
            % \\
            % russian
            ru & 27
            \\
            % German
            de &  28
            % both better than ALL sources
            \\
            % Hebrew
            he & 29
            \\
            % czech
            cs &  27
            \\
            % romanian
            % ro &  
            % \\
            % slovak
            sk &  27
            \\
            % Serbian
            sr & 27
            \\
            % indonesian
            id & 26
            \\
            % finnish
            fi & 27  
            \\
            % chinese
            % zh &   
            % \\
            % arabic
            ar & 30
            \\
            % Korean
            ko & 27
            \\
            % hindi
            hi &  27
            \\
            % Japanese
            ja &  29
            \\
            % turkish
            %  tr &   
            % \\
            % eu &  
            % \\
            fa &  27
            \\
            % \hline
            % {\bf Average} & - & 87.17 & 87.21 & {\bf 87.62 }
            % \\
		%===========
		\hline
	\end{tabular}

	\caption{
	\label{tab:num-source-select-lang-udpos}
{Number of sources selected from 30 different languages by \mym  for the task of cross-lingual POS tagging. 
}
		}
\vspace{-18pt}
\end{table}

\cleardoublepage

\clearpage

\end{document}